\documentclass[10pt,a4paper]{article}
\usepackage[scale=.8]{geometry}
\usepackage[affil-it]{authblk}
\usepackage{amssymb,amsmath,amsthm,amsbsy}
\usepackage{physics} 
\usepackage[pdftex,bookmarksnumbered]{hyperref}
\usepackage{graphicx}
\usepackage{epstopdf}
\usepackage{epsfig}
\usepackage{multicol}
\usepackage{subfigure}
\usepackage{multirow}
\usepackage{algorithm}
\usepackage{algpseudocode}

\usepackage{setspace}
\usepackage{color}
\usepackage[dvipsnames]{xcolor}
\usepackage{lineno}
\usepackage{diagbox}
\usepackage{rotating}

\usepackage{tcolorbox}
\usepackage[width=\textwidth]{caption}
\usepackage{lscape}
\usepackage{booktabs}
\usepackage{adjustbox}
\usepackage{mathtools}
\usepackage{bm}
\usepackage{lineno}
\usepackage{adjustbox}
\usepackage{lipsum}
\usepackage{anyfontsize}
\usepackage{xcolor}

\title{\Large\textbf{PDE-constrained Gaussian process surrogate modeling with uncertain data locations}}
\author{Dongwei Ye$^{1,}$\thanks{Corresponding author (e-mail: \texttt{d.ye-1@utwente.nl}).}, Weihao Yan$^1$, Christoph Brune$^1$, Mengwu Guo$^{1,2}$}
\affil{$^1$Department of Applied Mathematics, University of Twente, the Netherlands \\ \smallskip
$^2$Centre for Mathematical Sciences, Lund University, Sweden}
\date{}

\begin{document}

\maketitle 
\thispagestyle{empty}

\noindent\textit{Abstract}: Gaussian process regression is widely applied in computational science and engineering for surrogate modeling owning to its kernel-based and probabilistic nature. In this work, we propose a Bayesian approach that integrates the variability of input data into the Gaussian process regression for function and partial differential equation approximation. Leveraging two types of observables -- noise-corrupted outputs with certain inputs and those with prior-distribution-defined uncertain inputs, a posterior distribution of uncertain inputs is estimated via Bayesian inference. Thereafter, such quantified uncertainties of inputs are incorporated into Gaussian process predictions by means of marginalization. The setting of two types of data aligned with common scenarios of constructing surrogate models for the solutions of partial differential equations, where the data of boundary conditions and initial conditions are typically known while the data of solution may involve uncertainties due to the measurement or stochasticity. The effectiveness of the proposed method is demonstrated through several numerical examples including multiple one-dimensional functions, the heat equation and Allen-Cahn equation. A consistently good performance of generalization is observed, and a substantial reduction in the predictive uncertainties is achieved by the Bayesian inference of uncertain inputs.

\vspace{1mm}

\section{Introduction}
Data-driven methods have been extensively developed to serve as an alternative for approximating the solution of ordinary/partial differential equations (O/PDEs) in computational science and engineering \cite{Cuomo2022,arzani2021data,sanderse2024scientific,brunton2024promising}. Those models offer accurate prediction by learning the latent information embedded in the training data. Some of them also incorporate physics constraints to avoid nonphysical artifacts and further improve the generalization of predictive performance. Their efficiency and effectiveness have been showcased in a wide variety of applications \cite{phellan2021real,bock2019review,brunton2020machine}.

Uncertainty quantification is an essential part of the model validation process as modeling may inevitably involve incertitude due to the lack of knowledge and/or intrinsic system variability. Most of the uncertainties of physics-based modeling lie in model assumptions, model parameters and initial/boundary conditions, while the uncertainties in data-driven modeling mainly originate from the data for training, which are possibly corrupted by the numerical discrepancy, measurement error or stochasticity. How to properly quantify and reduce the uncertainties associated with a data-driven model leveraging the \textit{a priori} knowledge remains one of the important topics in scientific machine learning. 

Gaussian process (GP) regression is one of the state-of-the-art methods for data-driven modeling \cite{williams2006gaussian}. GP was first proposed by Krige for geo-statistical analysis \cite{Krige1951} and has been extensively studied to address interpolation and regression problems \cite{Chang2015, Gulian2019, costabal2019multi, Ye2022}. It is frequently synergized with Bayesian inference owing to its probabilistic nature \cite{Frigola2013, Alaa2017}. GP prior over the function to approximate is defined by its mean and covariance functions. The kernel representation of the covariance reflects the correlation between data and can also be viewed as a feature map to the reproducing kernel Hilbert space \cite{williams2006gaussian}. Conditioning on the training data, GP posterior offers a distribution of the approximated function, which includes the assessment of predictive uncertainty. GP can also serve as a surrogate model to approximate PDEs by exploiting the kernel structure to embed physics constraints. It was first proposed by Alvarez et al. \cite{alvarez2009latent} where the GP was employed to uncover the correlation between the state of a dynamical system and its latent force through linear operations. The method was subsequently extended to characterize the correlation between solution and external source term through differential operator in linear and nonlinear O/PDEs \cite{raissi2018hidden, besginow2022constraining, sarkka2011linear, ye2024gaussian, raissi2018numerical, chen2021solving, batlle2023error}. 

\begin{figure}
    \centering
    \includegraphics[width=0.65\linewidth]{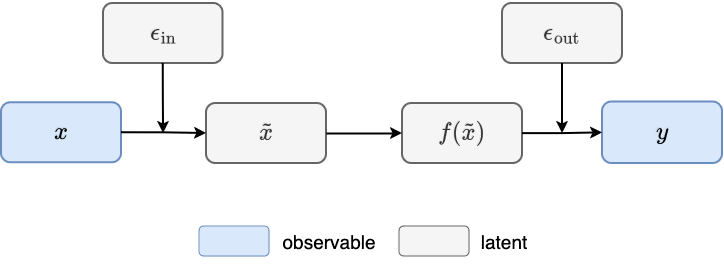}
    \caption{A schematic diagram of input data location corrupted by noise. The blue boxes and gray boxes denote observable and latent variables respectively.}
    \label{fig:three_cases}
\end{figure}

Most GP models are established based on the assumption that the input data have full certainty, while the data of model response are possibly corrupted by measurement noise and/or perturbed by intrinsic system stochasticity. GP regression models such uncertainties with an additive white Gaussian noise term, whose variance is calibrated as a part of hyperparameters. However, it is often the case that the assumption of uncertainties exclusively on output data is too strong and it is crucial to take uncertain data locations (inputs) into account as well. An intuitive example is the sensor placement and data measurement for the predictive monitoring of engineering assets, in which case sensors are deployed at several particular positions to examine certain mechanical behaviors of interest. Due to possible human errors, structure deformation, or interference of the ambient environment, the actual sensor locations could deviate from the desired placement, and the observational data are consequently measured from uncertain input locations. Similar circumstances also occur in mapping problems of robotic navigation \cite{Ghaffari2007}, system identification in marine science \cite{BURKHART2014189}. A schematic diagram of uncertain input location is shown in Figure~\ref{fig:three_cases}.

Existing methods on GP regression with uncertain input data can be categorized into two types, and both of them aim to find a robust way to quantify the propagation of uncertainties from data locations to predictions to guarantee a good generalization performance. The first type reflects the input uncertainties in GP kernel functions. In \cite{Girard2004}, a second-order Taylor expansion was performed to correct the covariance function with uncertain inputs, and the incertitude in both input and output data can be learned via maximum likelihood. \cite{Dallaire2009,DALLAIRE20111945} directly marginalized the distribution of data inputs in the covariance functions, which presented better generality than \cite{Girard2004} as the high-order terms were not omitted. 
The second type of methods assumes that the input uncertainties can be viewed as an additional noise term on the output, which aligns with the form of heterogeneous GP \cite{Mchutchon2011}. This method shows the intuition that the input uncertainties have a significant influence on the area where the output changes rapidly. 
An alternative is to fuse the determination of data locations into hyperparameter estimation \cite{Quionero2003}, for which gradient descent is used to optimize evidence lower bound integrated with the input distribution. 

In this work, we propose a new method that integrates input variability into GP regression, aided by a Bayesian inference of uncertain data locations. We assume the availability of two types of data: noise-corrupted outputs with certain inputs, and those with uncertain input locations described by a prior distribution. Through the Bayes' rule, the proposed method estimates a posterior distribution of the uncertain input locations by leveraging all the available information from the data and thereafter integrates such quantified uncertainties into GP predictions through marginalization. The setting of two types of data aligned with common scenarios of constructing surrogate model for PDE. Typically PDEs themselves and their boundary and initial conditions are known exactly while the data of solution may involve uncertainties due to the discrepancy from measurement or stochasticity. The proposed method therefore can be applied to those scenarios and contributes to the reduction of uncertainty in both input and output data. 

The paper is arranged as follows. The proposed Bayesian approach for data location uncertainty is detailed in Section~\ref{sec:BayApproch} in the context of the regression for one-dimensional functions, while its application to PDEs is demonstrated in Section~\ref{sec:PDElearning}. Several numerical examples are presented in Section~\ref{sec:examples}, followed by the conclusion in Section~\ref{sec:conclusion}.

\section{Bayesian approach to GP regression
with uncertain inputs}\label{sec:BayApproch}
\subsection{Gaussian process regression}
A general description of GP regression is presented in this section. Consider that the model response $y \in \mathbb{R}$ is measured from a physical process or a computer experiment as $y = f(\bm{x}) + \epsilon_n$, given the corresponding input $\bm{x}\in \mathbb{R}^d$. Here the regression function $f$ follows a GP prior with a mean function, and a covariance (kernel) function $k(\cdot,\cdot)$. The mean function is set to zero for simplicity in this work. The covariance function is featured with hyperparameters $\bm{\theta}$, i.e., $f(\cdot)\sim \mathcal{GP}(0,k(\cdot\,,\cdot\,;{\bm{\theta}}))$, and $\epsilon_n\sim \mathcal{N} (0,\sigma_n^2)$ denotes an independent white noise term.  Given the observed data collection $\ (\mathbf{X},\mathbf{y}) = \{(\bm{x}_i,y_i)\}^{N}_{i=1}$, 
a joint normal distribution between the observed data and the output $f(\bm{x}^*)$ at a test point $\bm{x}^*$ can be hence defined by the GP prior as
\begin{equation}
    \left[\begin{array}{c} \mathbf{y} \\ f(\bm{x}^*) \end{array}\right] \Big|~\mathbf{X},\bm{\theta},\sigma_n^2
\sim 
\mathcal{N}\left(\mathbf{0},\left[\begin{array}{cc}
k( \mathbf{X}, \mathbf{X};{\bm{\theta}})+\sigma_{n}^{2} \mathbf{I} & k\left(\mathbf{X},\bm{x}^{*};{\bm{\theta}}\right) \\
k\left(\bm{x}^{*}, \mathbf{X};{\bm{\theta}}\right) & k\left(\bm{x}^{*}, \bm{x}^{*};{\bm{\theta}}\right)
\end{array}\right]\right).
\end{equation}
Note that the choice of kernel should reflect the desired properties of the regression function. The hyperparameters $\bm{\theta}$ in the kernel as well as the noise variance $\sigma_n^2$ are often determined by maximizing the log-marginal likelihood:
\begin{equation}\label{eq:marginal_likelihood}
\begin{split}
    (\Tilde{\bm{\theta}},\Tilde{\sigma}_n^{2})  & =\arg \max _{\boldsymbol{\theta}, \sigma_n^2}~ \log p(\mathbf{y}| \mathbf{X}, \boldsymbol{\theta},\sigma^2_n) \\
    & = \arg \max _{\boldsymbol{\theta}, \sigma_n^2} \Big[ -\frac{1}{2} \mathbf{y}^{\top}\left( \bm{\mathrm{K}}(\boldsymbol{\theta}) +\sigma_{n}^{2} \mathbf{I}\right)^{-1} \mathbf{y} -\frac{1}{2} \log \left| \bm{\mathrm{K}}(\boldsymbol{\theta})+\sigma_{n}^{2}\mathbf{I}\right|-\frac{N}{2} \log 2 \pi  \Big],
\end{split}
\end{equation}
in which $\bm{\mathrm{K}}(\bm{\theta}) := k(\mathbf{X},\mathbf{X};{\bm{\theta}})$. 
Conditioning on the observed data, the posterior predictive distribution of the noise-free output $f(\bm{x}^*)$ at a test point $\bm{x}^*$ still follows a normal distribution:
\begin{equation}\label{gppredictive}
    f(\bm{x}^*) | \mathbf{X},\mathbf{y},\Tilde{\bm{\theta}},\Tilde{\sigma}_n^{2} \sim \mathcal{N}\Bigl( m(\bm{x}^*|\mathbf{X},\mathbf{y}),v(\bm{x}^*|\mathbf{X})\Bigr),
\end{equation}
where 
\begin{equation*}
    m(\bm{x}^*|\mathbf{X},\mathbf{y})= k(\bm{x}^*,\mathbf{X};{\Tilde{\bm{\theta}}})[\bm{\mathrm{K}}(\Tilde{\bm{\theta}})+\Tilde{\sigma}_n^{2} \mathbf{I}]^{-1}\mathbf{y},
\end{equation*}
\begin{equation*}
    v(\bm{x}^*|\mathbf{X}) = k(\bm{x}^*,\bm{x}^*;{\Tilde{\bm{\theta}}}) -k(\bm{x}^*,\mathbf{X};{\Tilde{\bm{\theta}}}) \big[\bm{\mathrm{K}}(\Tilde{\bm{\theta}})+\Tilde{\sigma}_n^{2}\mathbf{I}\big]^{-1} k(\mathbf{X},\bm{x}^*;{\Tilde{\bm{\theta}}}).
\end{equation*}
stand for the predictive mean and corresponding predictive variance respectively.

\subsection{Prior information of uncertain data locations}
To quantify the uncertainties in the input location, two types of training data are considered: $(\mathbf{X}^c,\mathbf{y}^c) = \{(\bm{x}^c_i,y^c_i)\}_{i=1}^{N_c}$ with certain input locations $\bm{x}_i^c \in \mathbb{R}^{d}$ and their corresponding observations $y_i^c \in \mathbb{R}$, and $(\mathbf{X}^u,\mathbf{y}^u)=\{(\bm{x}^u_i,y^u_i)\}_{i=1}^{N_u}$ consisting of uncertain input locations $\bm{x}_i^u $ and their corresponding observations $y_i^u$. Each uncertain input $\bm{x}^u_i$ is considered as a random variable associated with an independent prior following a normal distribution,
\begin{equation}
    \bm{x}^u_i \sim \mathcal{N}(\bm{\mu}_i,\text{diag}(s_{i,1}^2,s_{i,2}^2,\dots,s_{i,d}^2)), \hspace{1cm}  
    i = 1,2,\dots,N_u,
\end{equation}
where $\bm{\mu}_i$ denotes the mean vector of the prior; the uncertain inputs are additionally assumed to have uncorrelated entries of coordinates, and the covariance matrix is thus written as a diagonal matrix $\text{diag}(s_{i,1}^2,s_{i,2}^2,\dots,s_{i,d}^2)$ with $s_{i,j}^2$ being the variance of the $j$th coordinate of the $i$th uncertain data location. Therefore, the joint prior distribution of all the uncertain inputs is given by:
\begin{equation}
    p(\mathbf{X}^u) = \prod_{i=1}^{N_u}p(\bm{x}^u_i|\bm{\mu}_i,s_{i,1}^2,s_{i,2}^2,\dots,s_{i,d}^2).
\end{equation}
where $\{\bm{\mu}_i,s_{i,1}^2,s_{i,2}^2,\dots,s_{i,d}^2\}^{N_u}_{i=1}$ collects all the means and variances defining Gaussian prior of the uncertain inputs. In this work, we predefine those means and variances to let $\bm{\mu}_i$ represent an initial guess of the uncertain input location and $\pm 2s_{i,j}$ reflect the 95\% confidence.

\subsection{Bayesian inference of uncertain data locations}
The uncertain input locations can be inferred in a Bayesian manner from the combination of their prior distribution and corresponding outputs, aided by the GP assumption for the regression function. To be subsequently applied for predictions, the posterior distribution of these uncertain inputs is directly given by Bayes' rule as follows:
\begin{equation}\label{eq:posterior_input}
    p({\mathbf{X}^u} |\, \mathbf{X}^c, \mathbf{y}^u, \mathbf{y}^c, \bm{\theta},\sigma_n^2) \propto p({\mathbf{X}^u}) ~ p(\mathbf{y}^u,\mathbf{y}^c |\, \mathbf{X}^u,\mathbf{X}^c,\bm{\theta},\sigma_n^2)\,,
\end{equation}
where the second term is a likelihood function defined by the GP, i.e., 
\begin{equation} \label{eq:likelihood}
        \left[\begin{array}{c} \mathbf{y}^c \\ \mathbf{y}^u \end{array}\right] \Big| ~\mathbf{X}^u,\mathbf{X}^c,\bm{\theta},\sigma_n^2
\sim 
\mathcal{N}\left(\mathbf{0},\left[\begin{array}{cc}
k( \mathbf{X}^c, \mathbf{X}^c;{\bm{\theta}})+\sigma_{n}^{2} \mathbf{I} & k\left(\mathbf{X}^c,\mathbf{X}^{u};{\bm{\theta}}\right) \\
k\left(\mathbf{X}^{u}, \mathbf{X}^c;{\bm{\theta}}\right) & k\left(\mathbf{X}^{u}, \mathbf{X}^{u};{\bm{\theta}}\right)+\sigma_{n}^{2} \mathbf{I}
\end{array}\right]\right).
\end{equation}

This inference of Equation~\eqref{eq:posterior_input} can barely be evaluated exactly as its marginal likelihood, given as
\begin{equation}
    p(\mathbf{y}^u,\mathbf{y}^c |\, \mathbf{X}^c,\bm{\theta},\sigma_n^2) = \int p(\mathbf{X}^u) p(\mathbf{y}^u,\mathbf{y}^c |\, \mathbf{X}^u,\mathbf{X}^c, \bm{\theta},\sigma_n^2) ~\textrm{d}\mathbf{X}^u\,,
\end{equation}
is typically intractable to compute in practice. Therefore, the Markov Chain Monte Carlo (MCMC), variational inference or other Bayesian inference techniques can be applied to fetch the posterior samples or to approximate the distribution of uncertain data locations.

To determine the optimal values of GP hyperparameters $(\bm{\theta},\sigma_n^2)$ in \eqref{eq:posterior_input}, we maximize an approximation of the possibly intractable marginal likelihood. If there are sufficient amount of certain data available, the hyperparameters can be evaluated exclusively on those data,
\begin{equation}\label{eq:hyperpara_esti_without_uncertain}
(\tilde{\bm{\theta}},\tilde{\sigma}_n^2)=\arg\max_{(\bm{\theta},\sigma_n^2)}\log p\left(\mathbf{y}^c |\, \mathbf{X}^c,\bm{\theta},\sigma_n^2\right)\,,
\end{equation}
whose specific formulation is given by \eqref{eq:marginal_likelihood}. The hyperparameters represent the information from certain data and can be leveraged to reduce the uncertainty in other data. Alternatively, the hyperparameters can be achieved from,
\begin{equation}\label{eq:hyperpara_esti}
(\tilde{\bm{\theta}},\tilde{\sigma}_n^2)=\arg\max_{(\bm{\theta},\sigma_n^2)}\log p\left(\mathbf{y}^u,\mathbf{y}^c |\,  \{\boldsymbol{\mu}_i\}_{i=1}^{N_u} , \mathbf{X}^c,\bm{\theta},\sigma_n^2\right)\,,
\end{equation}
if only a few certain data are available and outnumbered by uncertain data. Such evaluation includes the prior means of uncertain input locations, by which we include the information from the uncertain data. The GP hyperparameters can be subsequently marginalized by considering
\begin{equation}
\begin{split}
    & \int p(\cdot |\, \mathbf{X}^c, \mathbf{y}^u, \mathbf{y}^c, \bm{\theta},\sigma_n^2) p(\bm{\theta},\sigma_n^2 |\, \mathbf{X}^c, \mathbf{y}^u, \mathbf{y}^c)~\textrm{d}\bm{\theta}~\textrm{d}\sigma_n^2\\
    = ~& \int p(\cdot |\, \mathbf{X}^c, \mathbf{y}^u, \mathbf{y}^c, \bm{\theta},\sigma_n^2) \delta(\bm{\theta}-\tilde{\bm{\theta}},\sigma_n^2-\tilde{\sigma}_n^2)~\textrm{d}\bm{\theta}~\textrm{d}\sigma_n^2
    = ~p(\cdot |\, \mathbf{X}^c, \mathbf{y}^u, \mathbf{y}^c,\tilde{\bm{\theta}},\tilde{\sigma}_n^2)\,,
\end{split}
\end{equation}
in which $\delta$ denotes the Dirac-delta function. This marginalization applies to both \eqref{eq:posterior_input} and the Bayesian predictions in the next subsection.

\subsection{Bayesian predictions for Gaussian process regression}
Once the posterior distribution of the uncertain inputs is achieved, the predictive distribution at a test point $\bm{x}^*$ can be further computed by marginalizing out the posterior, i.e.,

\begin{align}
    & p(f(\bm{x}^*) |\, \mathbf{X}^c,\mathbf{y}^c,\mathbf{y}^u,\tilde{\bm{\theta}},\tilde{\sigma}_n^2) \notag\\
    = ~& \int p(f(\bm{x}^*) |\, \mathbf{X}^c,\mathbf{X}^u,\mathbf{y}^c,\mathbf{y}^u, \tilde{\bm{\theta}},\tilde{\sigma}_n^2) ~p(\mathbf{X}^u |\, \mathbf{X}^c,\mathbf{y}^u,\mathbf{y}^c,\tilde{\bm{\theta}},\tilde{\sigma}_n^2) ~\textrm{d}\mathbf{X}^u \label{eq:prediction_mar}
\end{align}
The samples extracted by MCMC can be reused in equation~\eqref{eq:prediction_mar} to evaluate the integral with Monte Carlo. On the other hand, $p(f(\bm{x}^*) |\, \mathbf{X}^c,\mathbf{X}^u,\mathbf{y}^c,\mathbf{y}^u, \tilde{\bm{\theta}},\tilde{\sigma}_n^2)$ corresponds to the predictive distribution \eqref{gppredictive} of a GP model conditioning on the combination of both certain data pairs and those with sampled locations of uncertain inputs. However, constructing an individual GP model from scratch for each posterior sample and performing prediction by conditioning on both certain and uncertain data simultaneously could lead to high and redundant computational costs. An alternative way to perform marginalization is to separate the conditioning procedure for certain data and uncertain samples. A GP based only on the certain data $(\mathbf{X}^c,\mathbf{y}^c)$ can be constructed first, i.e.,
\begin{equation}\label{eq:first_condition}
    p(f(\bm{x}) | \mathbf{X}^c, \mathbf{y}^c) \sim \mathcal{GP}(m^c(\bm{x}),k^c(\bm{x},\bm{x}'))
\end{equation}
where 
\begin{equation*}
    m^c(\bm{x}) = k(\bm{x},\mathbf{X}^c) [k(\mathbf{X}^c,\mathbf{X}^c) + \tilde{\sigma}_n^2\mathbf{I}]^{-1}\mathbf{y}^c
\end{equation*}
\begin{equation*}
   k^c(\bm{x},\bm{x}') =  k(\bm{x},\bm{x}') - [k(\bm{x},\mathbf{X}^c) k(\mathbf{X}^c,\mathbf{X}^c)+\tilde{\sigma}_n^2\mathbf{I}]^{-1}k(\mathbf{X}^c,\bm{x}'). 
\end{equation*}    
Therefore, $p(f(\bm{x}^*) |\, \mathbf{X}^c,\mathbf{X}^u,\mathbf{y}^c,\mathbf{y}^u, \tilde{\bm{\theta}},\tilde{\sigma}_n^2)$ can be subsequently computed by leveraging the Equation~\eqref{eq:first_condition} as the new GP prior and performing another conditioning based on each uncertain samples of $\bm{x}^u_i$ sampled from their posterior distribution. Such a procedure avoids repeatedly computing the possible large covariance matrix of certain data and inversion operation for conditioning.
The first and second moments of the marginal predictive distribution \eqref{eq:prediction_mar} are derived as follows:

\begin{equation}
     \mathbb{E} \big[f(\bm{x}^*) |\, \mathbf{X}^c,\mathbf{y}^c,\mathbf{y}^u \big]  = \mathbb{E}_{\mathbf{X}^u |\, \mathbf{X}^c,\mathbf{y}^u,\mathbf{y}^c} \big[m(\bm{x}^* |\, \mathbf{X}^c\cup\mathbf{X}^u,\mathbf{y}^c \cup \mathbf{y}^u) \big]\,,
\end{equation}
and similarly 
\begin{equation}
    \begin{aligned}
 \mathbb{V}ar \big[f(\bm{x}^*)|\, \mathbf{X}^c,\mathbf{y}^c,\mathbf{y}^u \big] = ~& \mathbb{E}_{\mathbf{X}^u |\, \mathbf{X}^c,\mathbf{y}^u,\mathbf{y}^c} \big[v(\bm{x}^* |\, \mathbf{X}^c\cup\mathbf{X}^u) \big]\\
    +~& \mathbb{V}ar_{\mathbf{X}^u | \mathbf{X}^c,\mathbf{y}^u,\mathbf{y}^c} \big[m(\bm{x}^* |\, \mathbf{X}^c \cup\mathbf{X}^u,\mathbf{y}^c \cup\mathbf{y}^u) \big]\,,
    \end{aligned}
\end{equation}
where $m(\cdot)$ and $v(\cdot)$ denote the mean and variance of a GP prediction, 
$\mathbb{E}_{\mathbf{X}^u | \mathbf{X}^c,\mathbf{y}^u,\mathbf{y}^c}$ and $\mathbb{V}ar_{\mathbf{X}^u | \mathbf{X}^c,\mathbf{y}^u,\mathbf{y}^c}$ indicate the expectation and variance values over the posterior distribution of $\textbf{X}^u$, respectively. 

\section{Surrogate modeling for linear PDEs using GP with uncertain inputs}\label{sec:PDElearning}
GP regression can also be leveraged to construct the surrogate model for the PDE solution as proposed \cite{raissi2018hidden}. In this section, the proposed Bayesian approach with GP regression for uncertain input location is presented in the context of surrogate modeling for PDEs. We consider the governing equations including physical parameters, source term and initial/boundary conditions to be known, while some observations of the solution are corrupted by noise. The objective is to reduce the uncertainty in the solution data leveraging certain data from known information and the GP model that incorporates the physics constraints, meanwhile also training a surrogate model using GP to predict the quantities of interest.

\subsection{Surrogate modeling for linear PDEs using GP}\label{sec:learn_pde}
A linear PDE system can be expressed as:
\begin{equation}\label{eq:PDE}
    \mathcal{L}_{\bm{x}} z(\bm{x}) = g(\bm{x})
\end{equation}
where $\mathcal{L}_{\bm{x}}$ denotes the differential operator applying on the solution of the systems $z(\bm{x})$, and $g(\bm{x})$ represents the external source term. Note that in the scenario of PDE, input $\bm{x} \in \mathbb{R}^d$ typically refers to the spatial and temporal coordinates of the computational domain. Consider function $z(\bm{x})$ follows a GP prior with zero mean and covariance function $k_{zz}(\bm{x},\bm{x}';\bm{\theta})$, i.e. 
\begin{equation}\label{eq:PDE_GP}
    z(\bm{x}) \sim \mathcal{GP}\Bigl(0,k_{zz}(\bm{x},\bm{x}';\bm{\theta})\Bigr)
\end{equation}
where $\bm{\theta}$ denotes the hyperparameters of the kernel function $k_{zz}$. By using the property of GP, i.e. Gaussianity can be preserved through linear operation,
\begin{equation}\label{eq:source_GP}
    \mathcal{L}_{\bm{x}} z(\bm{x}) = g(\bm{x}) \sim \mathcal{GP}\Bigl(0,k_{gg}(\bm{x},\bm{x}';\bm{\theta})\Bigr)
\end{equation}     
where the kernel function $k_{gg}(\bm{x},\bm{x}';\bm{\theta}) = \mathcal{L}_{\bm{x}} \mathcal{L}_{\bm{x}'} k_{zz}(\bm{x},\bm{x}';\bm{\theta})$. Similarly, the covariance function between $z(\bm{x})$ and $g(\bm{x}')$ can be written as $k_{zg}(\bm{x},\bm{x}';\bm{\theta}) = \mathcal{L}_{\bm{x}'} k_{zz}(\bm{x},\bm{x}';\bm{\theta})$. With observed data $(\mathbf{X}_z,\mathbf{y}_z)=\big\{(\bm{\bm{x}}_{z,i}, y_{z,i})\big\}_{i=1}^{N_z}$ and $(\mathbf{X}_g,\mathbf{y}_g)=\big\{(\bm{x}_{g,i}, y_{g,i})\big\}_{i=1}^{N_g}$, obtained from measurement $y_z = z(\bm{x}_z)+\epsilon_z$ and $y_g = g(\bm{x}_g)+\epsilon_g$ respectively, where $\epsilon_z \sim  \mathcal{N}(0,\sigma_z^2)$ and $\epsilon_g \sim \mathcal{N}(0,\sigma_g^2)$. The hyperparameters $\bm{\theta}$ of the kernel as well as the noise variances ${\sigma}_z^2$ and ${\sigma}_g^2$ can be estimated via maximizing the log marginal likelihood:
\begin{equation}
\begin{aligned}
(\tilde{\bm{\theta}},\tilde{\sigma}_z^2,\tilde{\sigma}_g^2 )& = \arg\max_{\bm{\theta},\sigma_z^2,\sigma_g^2} \log p(\mathbf{y}_z.\mathbf{y}_g | \bm{\theta},\sigma_z^2,\sigma_g^2)\\
&= \frac{1}{2} \log |\bm{\mathrm{K}}| +\frac{1}{2}\mathbf{y}^{\top}\bm{\mathrm{K}}^{-1}\mathbf{y} +\frac{N_z+N_g}{2}\log 2\pi
\end{aligned}
\end{equation}
where
$$\mathbf{y} =
\left[\begin{array}{l}\mathbf{y}_z \\ \mathbf{y}_g
\end{array}\right],\quad \text{and}\quad 
\bm{\mathrm{K}}(\bm{\theta},\sigma_z^2,\sigma_g^2) = \left[\begin{array}{cc}
k_{zz}\left(\mathbf{X}_z, \mathbf{X}_z ; \bm{\theta}\right)+\sigma_{z}^2 \boldsymbol{I}_{z} & k_{z g}\left(\mathbf{X}_z, \mathbf{X}_g ; \bm{\theta}\right) \\
k_{gz}\left(\mathbf{X}_g, \mathbf{X}_z ; \bm{\theta} \right) & k_{gg}\left(\mathbf{X}_g, \mathbf{X}_g ; \bm{\theta} \right)+\sigma_{g}^2 \boldsymbol{I}_{g}
\end{array}\right].$$
The solution data $\mathbf{y}$ and the prediction $z(\bm{x}^*)$ at a new point $\bm{x}^*$ follow a joint distribution,
\begin{equation} 
        \left[\begin{array}{c} \mathbf{y} \\ z(\bm{x}^*) \end{array}\right] \Big| ~\mathbf{X}_z,\mathbf{X}_g,\tilde{\bm{\theta}},\tilde{\sigma}_{z}^2,\tilde{\sigma}_{g}^2
\sim 
\mathcal{N}\left(\mathbf{0}, \,\left[\begin{array}{cc}
 \bm{\mathrm{K}}(\tilde{\bm{\theta}},\tilde{\sigma}_z^2,\tilde{\sigma}_g^2) & \bm{\mathrm{q}}_z^{\top}{(\boldsymbol{x}^*)} \\
\bm{\mathrm{q}}_z{(\boldsymbol{x}^*)} &  k_{zz}(\bm{x}^*,\bm{x}^*;\tilde{\bm{\theta}})
\end{array}\right]\right).
\end{equation}
where $\bm{\mathrm{q}}_z {(\boldsymbol{x}^*)}=\left[k_{zz}\left( \bm{x}^*,\mathbf{X}_z; \tilde{\bm{\theta}}\right) \, k_{zg}\left(\bm{x}^*, \mathbf{X}_g ;\tilde{\bm{\theta}} \right)\right]$. 
Therefore
\begin{equation} \label{eq:pde_prediction_u}
    p\Bigl( z(\bm{x}^*) | \, \mathbf{y},\mathbf{X}_z,\mathbf{X}_g,\tilde{\bm{\theta}},\tilde{\sigma}_{z}^2,\tilde{\sigma}_{g}^2 \Bigr) = \mathcal{N}\Bigl(\bar{z}(\bm{x}^*),v_z(\bm{x}^*)\Bigr)
\end{equation}
denotes the predictive distribution of the solution at $\bm{x}^*$, with mean $\bar{z}(\bm{x}^*) = \bm{\mathrm{q}}_z{(\boldsymbol{x}^*)} \bm{\mathrm{K}}^{-1} \mathbf{y}$ and variance $ v_z(\bm{x}^*)=k_{zz}(\bm{x}^*, \bm{x}^*)-\bm{\mathrm{q}}_z{(\boldsymbol{x}^*)} \bm{\mathrm{K}}^{-1} \bm{\mathrm{q}}_z^{\top}{(\boldsymbol{x}^*)}$. Note that here we drop hyperparameters and noise terms in the expression of predictive means and variance for simplicity.

\subsection{Surrogate modeling for linear PDEs using GP with uncertain inputs}
GP model can encode the information of PDE into its kernel, by which the correlation between the source term $g(\bm{x})$ and the solution $z(\bm{x})$ is constructed. Therefore the data from the source term can be considered as additional physics constraints that help the GP to learn and predict the quantity of interest. Generally, an unlimited amount of precise data of $g(\bm{x})$ can be generated synthetically since the source term of the PDE is fully known. Some data of $z(\bm{x})$ can also be evaluated precisely given well-defined initial and boundary conditions. However the other data of $z(\bm{x})$ are typically measured from experiments/sensors and are very likely corrupted by noise due to the measurement error or disruption. Therefore, we consider three types of data in this scenario:
\begin{itemize}
    \item $(\mathbf{X}_z^c,\mathbf{y}_z^c)=\big\{ (\bm{x}_{z,i}^c, y_{z,i}^c)  \big\}_{i=1}^{N_{zc}}$ denotes the solution data with certain locations, e.g. data of accurate measurement, boundary condition or initial conditions, where $y_{z,i}^c = z(\bm{x}_{z,i}^c)$.  
    \item   $(\mathbf{X}_g^c,\mathbf{y}_g^c)=\big\{ (\bm{x}_{g,i}^c,y_{g,i}^c) \big\}_{i=1}^{N_{gc}} $ denotes the synthetic data of source term $y_{g,i}^c = g(\bm{x}_{g,i}^c)$. Note that, both input and output data generated from initial/boundary condition and source term are noise-free.
    \item  $(\mathbf{X}_z^u,\mathbf{y}_z^u)=\big\{ (\bm{x}_{z,i}^u,y_{z,i}^u)  \big\}_{i=1}^{N_{zu}}$ denotes those noise-corrupted solution data, where $\bm{x}_{z,i}^u$ denote uncertain input locations and $y_{z,i}^u$ are corresponding observations. 
    
\end{itemize}
Similar to the scenario of a standard regression problem, each uncertain input location $\bm{x}_{z,i}^u$ is considered as a random variable with an independent prior distribution,
\begin{equation}
    \bm{x}^u_{z,i} \sim \mathcal{N}\big(\bm{\mu}_{z,i},\text{diag}(s_{i,1}^2,s_{i,2}^2,\dots,s_{i,d}^2)\big), \hspace{1cm}  
    i = 1,2,\dots,N_{zu},
\end{equation}
The joint prior distribution of all the uncertain input locations $\mathbf{X}^u_z$ can be written as,
\begin{equation}
    p(\mathbf{X}^u_z) = \prod_{i=1}^{N_{zu}}p(\bm{x}^u_{z,i}|\, \bm{\mu}_{z,i},s_{i,1}^2,s_{i,2}^2,\dots,s_{i,d}^2).
\end{equation}
where $\{ \bm{\mu}_{z,i}, s_{i,1}^2,s_{i,2}^2,\dots,s_{i,d}^2 \}_{i=1}^{N_{zu}}$ denotes the collection of all the prescribed means and variances of priors. The likelihood function can be formulated as a joint Gaussian distribution based on equation~\eqref{eq:PDE_GP} and \eqref{eq:source_GP},

\begin{equation}\label{eq:PDElikelihood}
\left[\begin{array}{c} \mathbf{y}_z^u \\ \mathbf{y}_z^c \\ \mathbf{y}_g^c \end{array}\right] \Big| ~\mathbf{X}_z^u,\mathbf{X}_z^c,\mathbf{X}_g^c
\sim 
\mathcal{N}\left(\mathbf{0},\left[\begin{array}{ccc}
k_{zz}( \mathbf{X}_z^u, \mathbf{X}_z^u) + \sigma^2_{z} \mathbf{I} & 
k_{zz}( \mathbf{X}_z^u, \mathbf{X}_z^c) &
k_{zg}( \mathbf{X}_z^u, \mathbf{X}_g^c)
\\
k_{zz}( \mathbf{X}_z^c, \mathbf{X}_z^u) &
k_{zz}( \mathbf{X}_z^c, \mathbf{X}_z^c) &
k_{zg}( \mathbf{X}_z^c, \mathbf{X}_g^c)\\

k_{gz}( \mathbf{X}_g^c, \mathbf{X}_z^u) &
k_{gz}( \mathbf{X}_g^c, \mathbf{X}_z^c) &
k_{gg}( \mathbf{X}_g^c, \mathbf{X}_g^c) 
\end{array}\right]\right).    
\end{equation}
For the simplicity of notation, we drop kernel hyperparameters $\bm{\theta}$ as well as noise variance $\sigma^2_z$ in \eqref{eq:PDElikelihood} as well as other expressions from now on. The kernel hyperparameters and noise variance are estimated via maximizing marginal likelihood as shown in Equation~\eqref{eq:hyperpara_esti} considering the sufficient availability of certain data from boundary/initial conditions and source term. The posterior distribution of uncertain input locations can be inferred via Bayes' rule:
\begin{equation}\label{eq:PDEposterior}
        p({\mathbf{X}^u_z} |\, \mathbf{X}_z^c,\mathbf{X}_g^c, \mathbf{y}_z^u,\mathbf{y}_z^c, \mathbf{y}_g^c) \propto p({\mathbf{X}^u} )~ p(\mathbf{y}_z^u,\mathbf{y}_z^c,\mathbf{y}_g^c | \, \mathbf{X}_z^u,\mathbf{X}_z^c,\mathbf{X}_g^c)
\end{equation}
MCMC, variational inference or other Bayesian inference techniques can be applied to approximate the posterior samples or distribution. The Bayesian prediction of $z(\bm{x}^*)$ is subsequently achieved by marginalizing out the posterior,
\begin{equation}\label{eq:PDEmarginalPred}
\begin{split}
& p\big(z(\bm{x}^*) |\, \mathbf{X}_z^c,\mathbf{X}_g^c,\mathbf{y}_z^u,\mathbf{y}_z^c,\mathbf{y}_g^c\big) \\
= ~& \int p\big(z(\bm{x}^*) | \, \mathbf{X}_z^u,\mathbf{X}_z^c,\mathbf{X}_g^c,\mathbf{y}_z^u,\mathbf{y}_z^c,\mathbf{y}_g^c\big) ~p\big({\mathbf{X}^u_z} | \, \mathbf{X}_z^c,\mathbf{X}_g^c, \mathbf{y}_z^u,\mathbf{y}_z^c, \mathbf{y}_g^c\big) ~\textrm{d}\mathbf{X}_z^u\,.
\end{split}
\end{equation}
We utilize MCMC samples to approximate the marginal predictive distribution of \eqref{eq:PDEmarginalPred}. The same technique as described in \eqref{eq:first_condition} can be applied to reduce the computational cost of marginalization by separating the conditioning operation of GP for certain data and uncertain data. To simplify the notation, we denotes $\mathbf{X}^c_{zg} =\{\mathbf{X}_z^c, \mathbf{X}_g^c\}$ as the collection data with certain location and  $\mathbf{y}^{cu}_{zg} = \{ \mathbf{y}_z^u, \mathbf{y}_z^c, \mathbf{y}_g^c \}$ as the collection of the output of all data.
The mean and variance of this distribution are expressed as,
\begin{equation}
\mathbb{E}\left[z^*| \boldsymbol{x}^*, \mathbf{X}^c_{zg}, \mathbf{y}^{cu}_{zg}\right]=\mathbb{E}_{\mathbf{X}_z^u | \mathbf{X}^c_{zg}, \mathbf{y}^{cu}_{zg}} \left[m \left(\boldsymbol{x}^*| \mathbf{X}^c_{zg} \cup \mathbf{X}_z^u,\mathbf{y}^{cu}_{zg} \right.\right.)]
\end{equation}
and
\begin{equation}
\begin{aligned}
\mathbb{V}ar\left[z^* | \boldsymbol{x}^*, \mathbf{X}^c_{zg}, \mathbf{y}^{cu}_{zg}\right] &=  \mathbb{E}_{\mathbf{X}_{z}^u | \mathbf{X}^c_{zg}, \mathbf{y}^{cu}_{zg}}\left[v\left(\boldsymbol{x}^* | \mathbf{X}^c_{zg} \cup \mathbf{X}_z^u\right)\right] \\
& + \mathbb{V}ar_{\mathbf{X}_z^u | \mathbf{X}^c_{zg}, \mathbf{y}^{cu}_{zg}}\left[m \left(\boldsymbol{x}^* | \mathbf{X}^c_{zg} \cup \mathbf{X}_z^u, \mathbf{y}^{cu}_{zg}\right.\right.)]
\end{aligned}
\end{equation}
Here, $m\left(\boldsymbol{x}^*| \mathbf{X}^c_{zg} \cup \mathbf{X}_z^u,\mathbf{y}^{cu}_{zg} \right.)$ and $v\left(\boldsymbol{x}^* | \mathbf{X}^c_{zg} \cup \mathbf{X}_z^u\right)$ represent the mean and variance of the predictive distribution of a GP model conditioning on the certain data together with the samples from the posterior distribution of uncertain inputs.

\section{Numerical Experiments}\label{sec:examples}
\subsection{Regression for 1D functions}
\begin{figure}[t!]
  \centering
  \includegraphics[width=\textwidth]{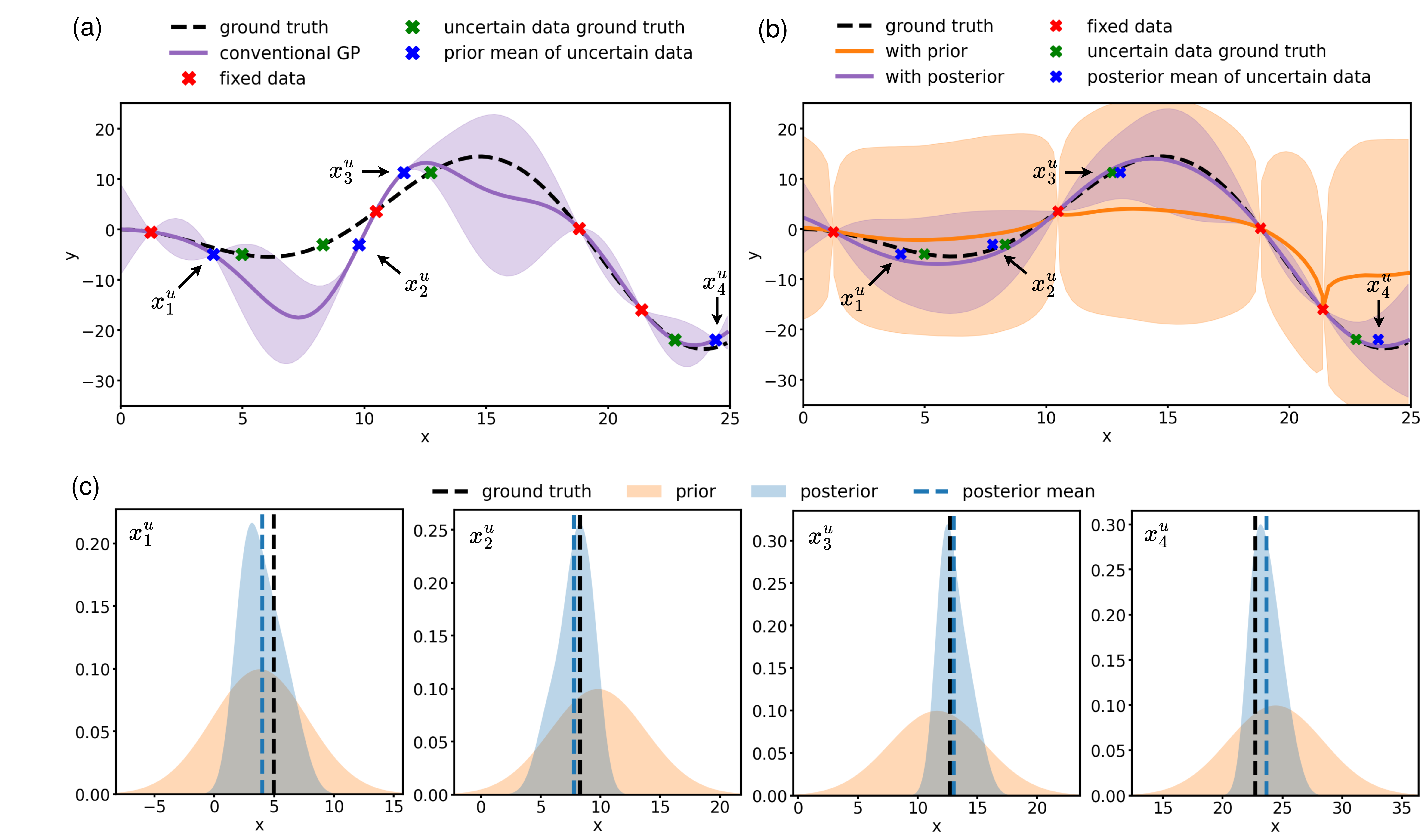}
  \caption{(a) Data points with certain and uncertain input locations generated from latent function $y=-x\sin(x/3)$, and the result of standard GP fitting; $x_1^u$ to $x_4^u$ denote the four uncertain input locations. (b) A comparison of GP predictions marginalized over the prior and posterior distributions of the uncertain input locations. (c) Prior and posterior marginal distributions of each uncertain input location. 
  Note that color bands in (a) and (b) represent the $\pm 2\sigma$ level of GP predictions in this figure. The predictions are depicted based on 100 test points distributed evenly over the domain of interest.
}
\label{fig:detailed_illustration}
\end{figure}

\begin{figure}[t]
  \centering
  \includegraphics[width=\textwidth]{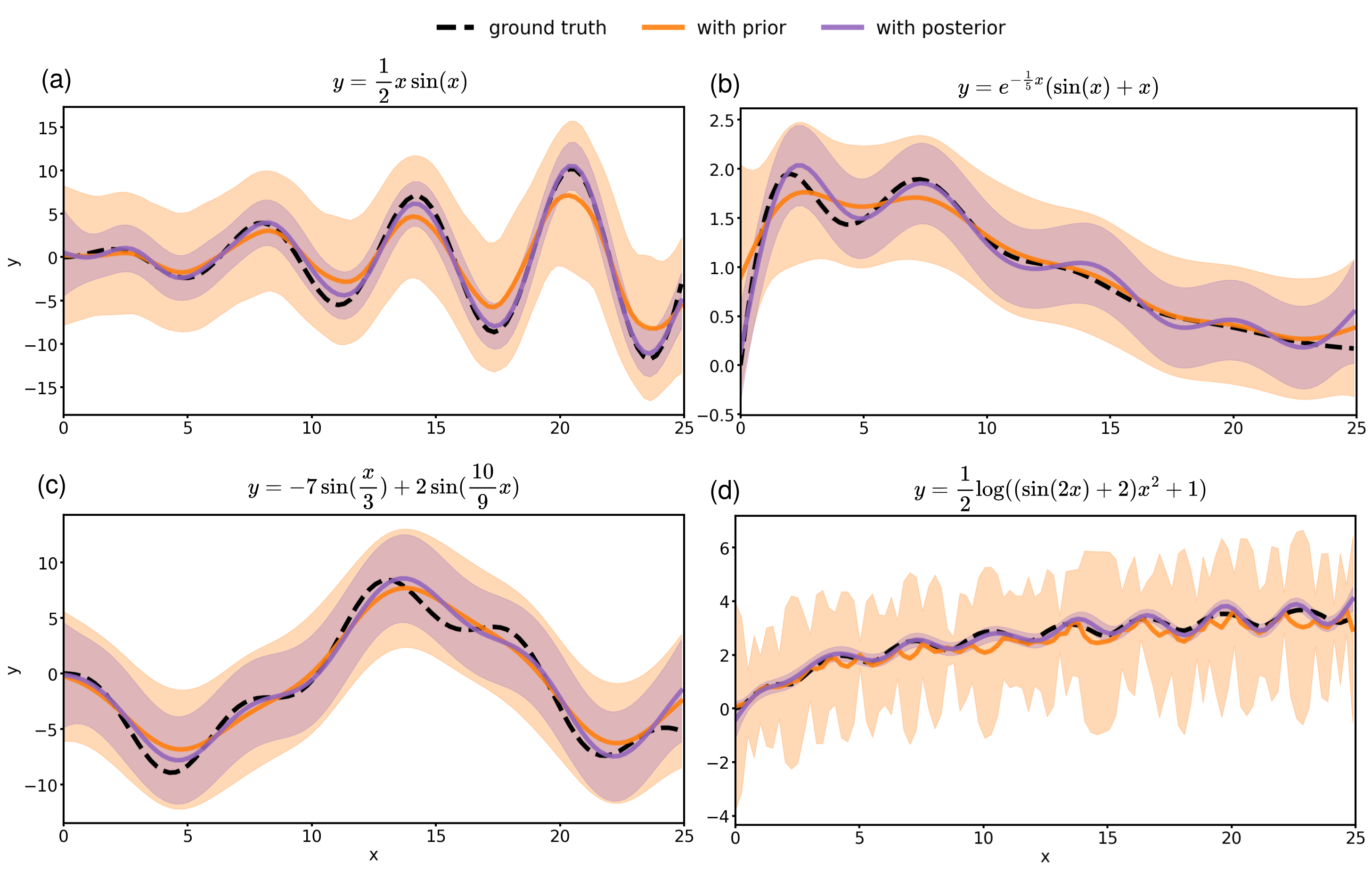}
  \caption{A comparison of GP predictions marginalized over the prior and posterior distributions of the uncertain input locations in four examples. Color bands represent the $\pm 2\sigma$ level of GP predictions in this figure. There are 30 training data points with certain inputs and another 30 with uncertain inputs, and the predictions are based on 100 test points distributed evenly over $[0,8\pi]$.}
  \label{fig:prediction_post}
\end{figure}

The application of the proposed method on multiple one-dimensional functions is demonstrated in this section. We use the mean squared error (MSE) to measure estimation errors of the uncertain input locations, and the mean squared prediction error (MSPE) over $N_\text{test}=100$ test points for GP predictions. These two error metrics are defined as follows,
\begin{equation}
    \text{MSE} = \frac{1}{N_u}\mathbb{E}_{\mathbf{X}_u}[\|\mathbf{X}_u-\hat{\mathbf{X}}_u\|^2]\,,\quad
    \text{MSPE} = \frac{1}{N_\text{test}}\mathbb{E}_{\mathbf{X}_u}\mathbb{E}_{f|\mathbf{X}_u}[\|f(\mathbf{X}^*)-\hat{f}(\mathbf{X}^*)\|^2]\,,\quad
\end{equation}
in which $\hat{\mathbf{X}}_u$ collects the actual locations of uncertain data inputs, $\mathbf{X}^*$ denotes the test points, $\hat{f}$ is the ground truth function for regression, and $\mathbf{X}_u$ may follow its prior \emph{or} posterior distribution.

To demonstrate the details of the proposed method, an example with eight data points is presented first. The training data are generated from a one-dimensional function $y = -x\sin(x/3)$. We assume that four of the data points have certain locations, while the other four are uncertain inputs $x_i^u$ ($i=1,2,3,4$), for which prior distributions are defined. The prior means of these uncertain data locations are computed by appending a perturbation to given locations, i.e., $\mu_i = \hat{x}_i^u + \epsilon$, where $\hat{x}_i^u$ denotes the actual location on the domain $[0,8\pi]$ and $\epsilon$ is sampled from a uniform distribution $\mathcal{U}(-2,2)$. The actual locations of these eight data points are sampled by the quasi-Monte Carlo method using a Sobol sequence \cite{Sobol1999}. 
The data points together with the ground truth of the function are shown in Figure~\ref{fig:detailed_illustration}(a). We are interested in the regression over $[0,8\pi]$ with an independent prior distribution $x^u_i \sim \mathcal{N}(\mu_i,s^2)$ for each uncertain input location. Assuming that we do not have much confidence in the priors, a relatively large variance $s^2=4$ is presumed. The measurement noise is omitted in this example for demonstration purposes. 

Figure~\ref{fig:detailed_illustration}(a) depicts the predictive distribution of a conventional GP model, which is trained with certain data inputs and prior means of uncertain inputs, paired with their corresponding observations.
It is evident that this GP model fails to incorporate the input uncertainties, and the mean and variance of the GP predictive function are misled by the perturbed locations of uncertain inputs in this interpolation task.

The results of the proposed Bayesian approach with GP regression is presented in Figures \ref{fig:detailed_illustration}(b) and (c). A posterior estimation of uncertain data locations is performed first through Bayesian inference. As the marginal likelihood (evidence) is computationally intractable, an MCMC sampling with the Metropolis algorithm is employed to directly fetch samples for the posterior distribution of uncertain inputs. The marginal prior (in orange) and posterior (in blue) distributions of each uncertain data location are presented in Figure~\ref{fig:detailed_illustration}(c). The distributions are approximated using samples and kernel density estimation \cite{Mats1982}. It is clear that the update from prior to posterior has driven the mean estimates of the uncertain inputs closer to the actual locations, meanwhile, the prior variances have been significantly reduced. 
It is worth noting that, in practice, the ground truth function is generally unknown, and thus it is typically impossible to claim if the inference has indeed improved our knowledge in uncertain input locations. Nevertheless, the Bayesian inference of these locations exploits all the available information from training data to reduce/update their uncertainties. 
Subsequently, the samples that MCMC fetches from the posterior are further used to construct the predictions for unseen locations over the entire domain $[0,8\pi]$. 
To show the reduction of uncertainty in GP regression as a result of Bayesian inference for uncertain inputs, we present in Figure~\ref{fig:detailed_illustration}(b) the GP predictions marginalized over the prior (in orange) and posterior (in purple) distributions of uncertain inputs. The former is affected by the significant uncertainties assumed in the prior and results in a high variance in the GP prediction. The mean function also deviates from the ground truth. On the contrary, the predictive mean with the posterior information achieved by the proposed method captures the latent function well, and its corresponding predictive variance demonstrates a good representation of the uncertain inputs, i.e., a relatively wide uncertainty band is observed around uncertain locations ($x_1^u$ to $x_4^u$), while the variance pinches to a small value at certain points. The MSPE at 100 test locations, respectively for the prior and posterior predictions by GP, are 143.2 and 11.9.

\begin{table}[t]
\renewcommand{\arraystretch}{1.3}
\begin{adjustbox}{width=1\textwidth}
\begin{tabular}{c|c|c|c|c|c|c}
\toprule
\multicolumn{1}{l|}{}& \multicolumn{3}{c|}{Uncertain input locations} & \multicolumn{3}{c}{GP predictions} \\
\midrule
 Numerical examples & \begin{tabular}[c]{@{}c@{}}prior vs\\ ground truth\end{tabular}   & \begin{tabular}[c]{@{}c@{}}posterior vs\\ ground truth\end{tabular} & \begin{tabular}[c]{@{}c@{}}relative\\ reduction \end{tabular}& \begin{tabular}[c]{@{}c@{}}prior vs\\ ground truth\end{tabular}&\begin{tabular}[c]{@{}c@{}}posterior vs\\ ground truth\end{tabular}& \begin{tabular}[c]{@{}c@{}}relative \\ reduction\end{tabular}\\
  \midrule
(a) $y = \frac{1}{2}x\sin(x)$& 1.10 & 0.45 & 59.1\% & 16.42& 2.33 & 85.8\%\\
(b) $y = e^{-\frac{1}{5}x}(\sin(x)+x)$& 1.10& 0.84& 23.6\% & 0.12& 0.05& 58.3\%\\
(c) $y = -7\sin(\frac{x}{3})+2\sin(\frac{10}{9}x)$& 1.10& 0.89& 19.1\% & 8.45& 4.92& 41.8\%\\
(d) $y = \frac{1}{2}\log((\sin(2x)+2)x^2+1)$& 1.10& 0.32& 70.9\%& 1.56& 0.05& 96.8\%\\
\bottomrule
\end{tabular}
\end{adjustbox}
\caption{A comparison of MSE for the uncertain inputs between prior and posterior distributions against ground truth locations, and that of MSPE between prior and posterior GP predictions against ground truth functions. The relative reduction refers to the percentage of corresponding error reduction from prior to posterior.}
\label{tab:MSPEcomparison}
\end{table}

The results of the other four one-dimensional functions are visualized in Figure~\ref{fig:prediction_post} and Table~\ref{tab:MSPEcomparison}. More data are utilized in these examples -- 30 each for certain and uncertain inputs, and output measurements are assumed to be corrupted by noise. 
Such existence of output noise plays a role in regularization, and the value of $\sigma_n^2$ can be fine-tuned to prevent over- and under-fitting in GP regression. Similarly, predictions with and without an update of input location inference are shown in Figure~\ref{fig:prediction_post}(a) for comparison. Both predictive mean functions have captured the pattern in the latent function; however, the one marginalized with the prior slightly underestimates at each wave peak and trough, while the other captures these values better. On the other hand, the $\pm 2\sigma$ level of the GP predictions, represented by the width of color bands, has been significantly reduced because of the update of uncertain input locations by Bayesian inference. In case (b) and (c), substantial improvement in the GP predictive distribution has been observed according to the comparison of MSPE in Table~\ref{tab:MSPEcomparison}. Moreover, the advantage of our proposed method in Figure~\ref{fig:prediction_post}(d) is evident. The result with prior information shows a poor predictive mean estimation, while our method captures the latent function well, and the narrow uncertainty band shows a high confidence in the prediction.
Among the four examples, in general, the proposed method with Bayesian inference for uncertain data locations consistently achieves a meaningful reduction in the predictive uncertainties and offers an improved predictive mean estimation for the GP regression. 

\subsection{Heat equation}
In this example, we present a heat equation problem using GP and leverage the existing information to improve the estimation of uncertain locations and to reduce their associated uncertainties. The heat equation describes the diffusion behavior of thermodynamics over time and the one-dimensional problem of which can be written as,
\begin{equation}
\begin{dcases}
    \frac{\partial z({x},t)}{\partial t} - \alpha \frac{\partial^2 z({x},t)}{\partial x^2} = g({x},t), & (x,t) \in (0,1)\times (0,1]\\
    g({x},t) = \left(16 \pi^2-1\right) e^{-t} \sin (4 \pi x),\\
    z(0,t) = z(1,t) = 0,\\
    z(x,0) = \mathrm{sin}(4\pi x),
\end{dcases}
\end{equation}
where $g({x},t)$ denotes the external source term;  $z({x},t)$ represents the temperature, and subject to the initial condition $\mathrm{sin}(4\pi x)$ and Dirichlet boundary condition $ z(0,t) = z(1,t) = 0$; Coefficient $\alpha$ denotes the diffusion coefficient and is set to be 1. The differential operator for the heat equation can be written as $\mathcal{L} = (\frac{\partial}{\partial t} - \alpha \frac{\partial^2}{\partial x^2})$ and the analytical solution of which $z({x},t)= e^{-t} \sin (4 \pi x)$ will be applied to measure the accuracy of GP prediction.

\begin{figure}[tb!]
    \centering
    \includegraphics[width=\linewidth]{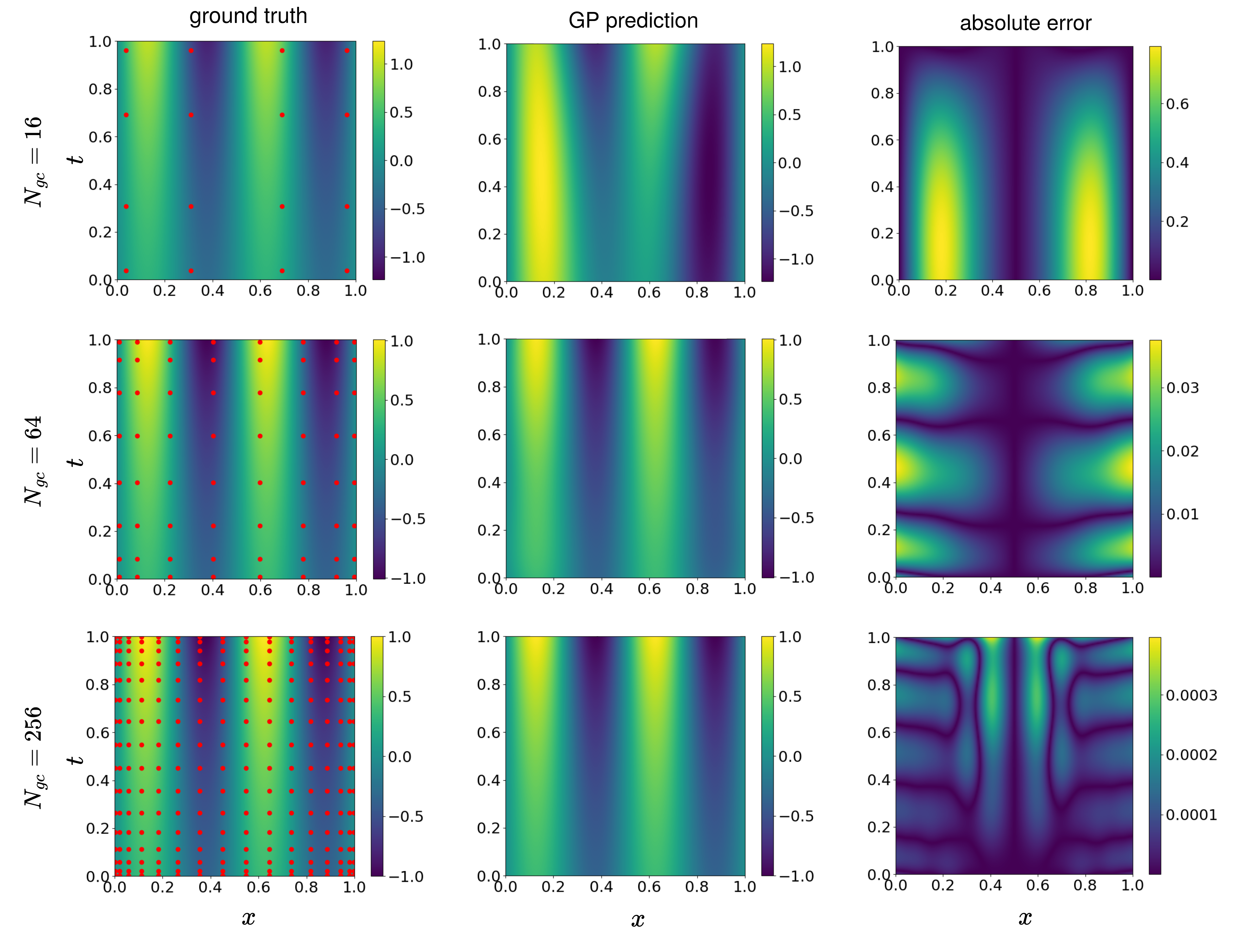}
    \caption{Comparisons of GP prediction and ground truth with training samples of the source term $N_{gc}= 16$, 64 and 256, and their corresponding absolute errors. The red dots on the first column represent the locations of training data of the source term generated from Chebyshev nodes.}
    \label{fig:HE_GP_check}
\end{figure}

\begin{figure}[th!]
    \centering
    \includegraphics[width=\linewidth]{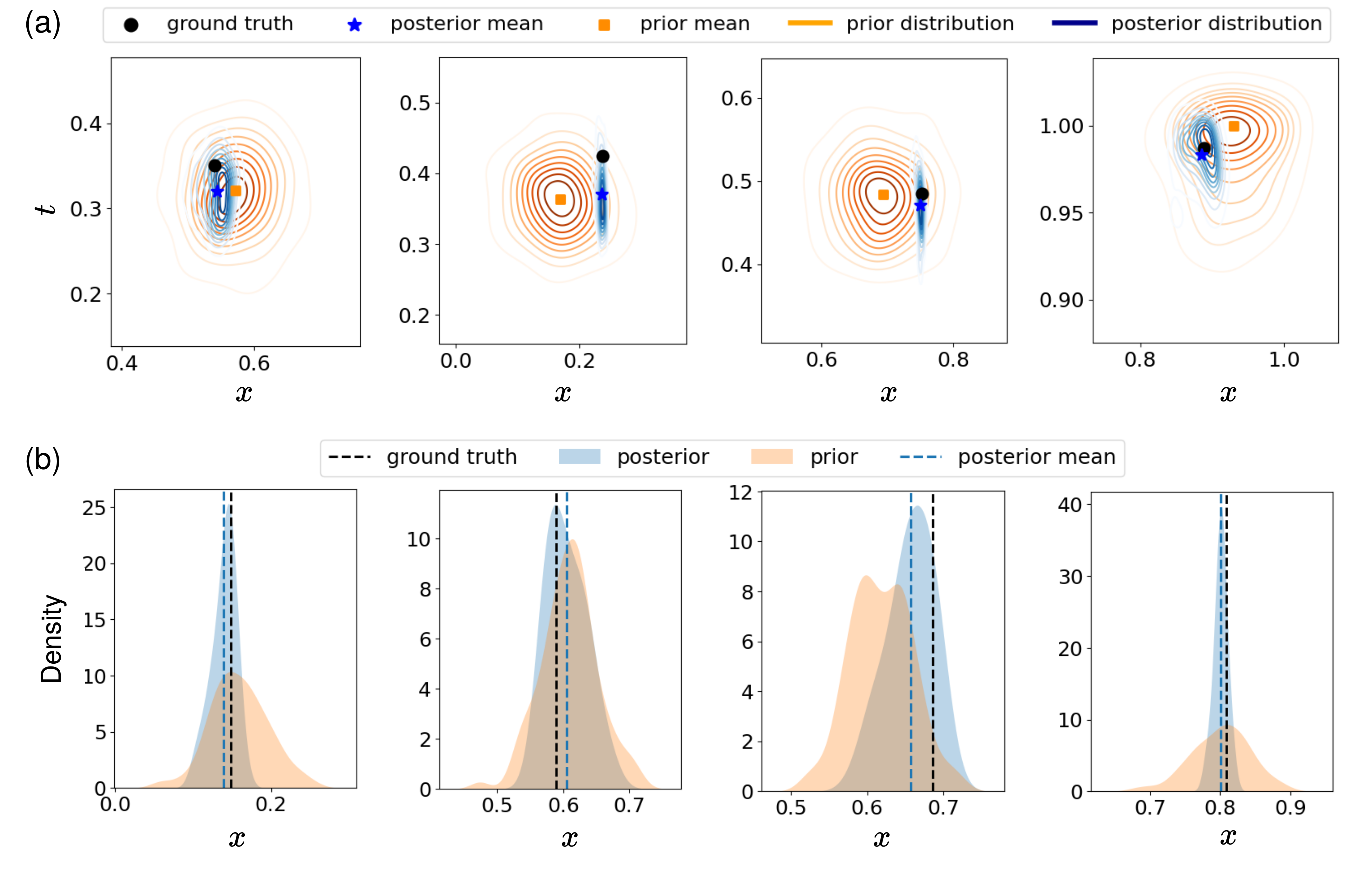}
    \caption{(a) Prior and posterior distributions of four data with uncertainties in location and time. The black dots denote the actual locations of data. Orange contours and squares denote their prior distributions and means. Blue contours and stars denote the posterior distributions and means. (b) Prior and posterior distributions of four data with uncertainties in location. The black dash line denotes the actual locations of data. Orange areas denote their prior distributions. Blue areas denote the posterior distribution and the blue dash line denotes their means.}
    \label{fig:HE_posterior}
\end{figure}

\begin{figure}[th!]
    \centering
    \includegraphics[width=\linewidth]{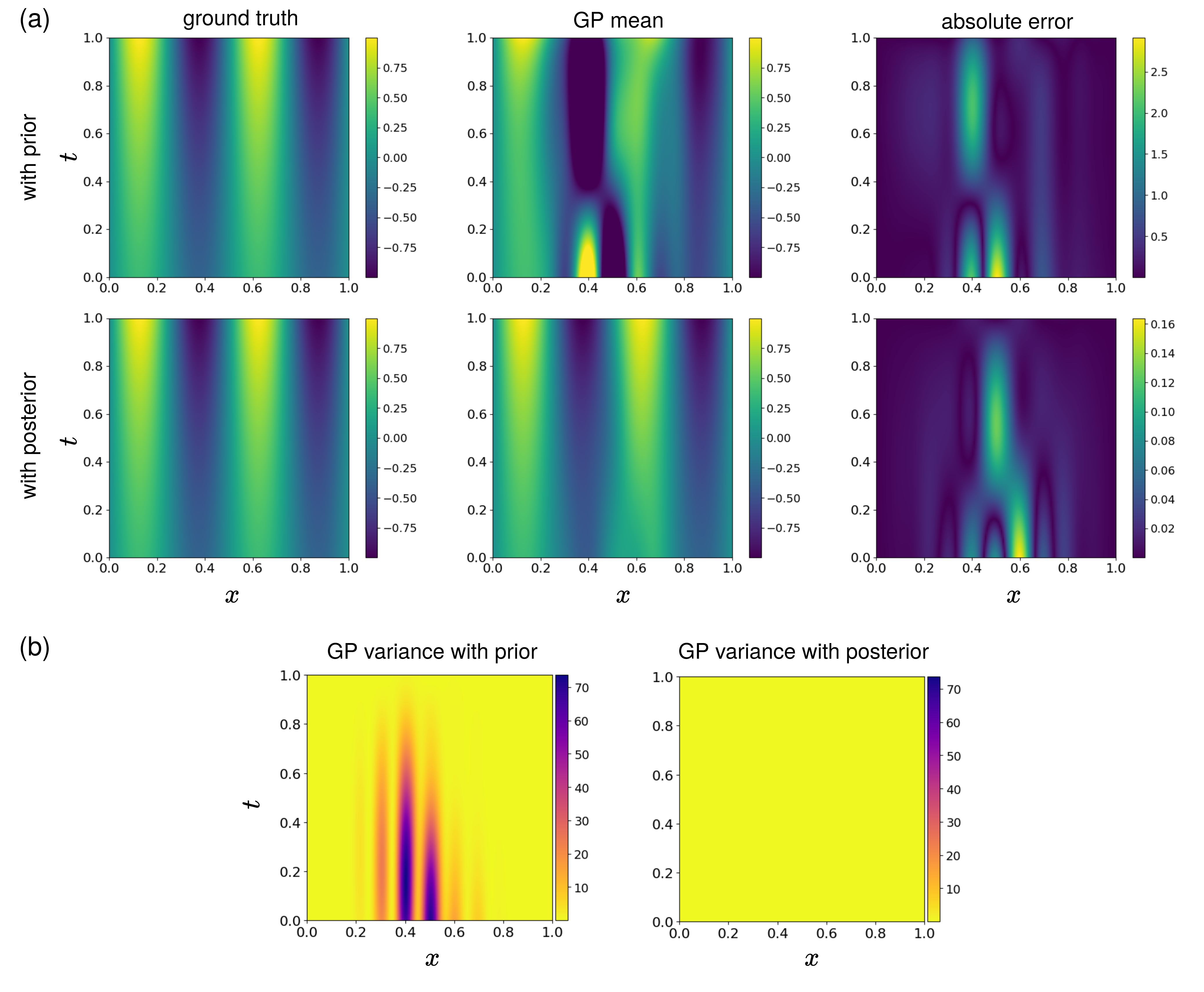}
    \caption{Means and variances of GP predictions combining with prior and posterior distributions of uncertainty data locations.}
    \label{fig:HE_GP_prediction}
\end{figure}

The data with certain locations contains the samples on the initial condition, boundary condition, and source term. Chebyshev nodes of the second kind are applied to sample the data of initial and boundary conditions, while the source term data are taken from Chebyshev nodes of the first kind. To ensure the GP surrogate model approximates the PDE solution sufficiently well, experiments have been conducted to assess its performance using varying sample sizes for the source term. The comparisons of GP predictions with 16, 64 and 256 samples against the ground truth solution of the heat equation are demonstrated in Figure~\ref{fig:HE_GP_check}. The red dots indicate the locations of the samples of the source term. The results show that the GP performance improves along with the increase in the sample size of the source term. The absolute error decreases around an order of magnitude when the samples increase from 16 to 64, and significantly decreases when the sample size reaches 256. Considering the decent performance achieved with 256 samples, the subsequent experiments for Bayesian inference and prediction are conducted based on this scenario.

Eight data locations with uncertainty are synthetically generated. Four of them have uncertainty in both $t$ and $x$, while the other four exhibit uncertainty in $x$ only. The prior means of these data are generated from their actual locations corrupted by a white noise that follows a normal distribution with zero mean and a standard deviation of 0.04. The standard deviation of prior distributions is also set to be 0.04. The posterior samples are subsequently obtained using the Metropolis algorithm. A comparison of the prior distributions and posterior distributions for uncertain data locations is demonstrated in Figure~\ref{fig:HE_posterior}. It is obvious that the variances of the posterior distributions have been significantly reduced for all the cases compared to the assumption given in prior. The information contained in the certain data contributes to variance reduction of uncertainty data location. Note that the variance reduction in the $x$ axis is much more evident than that for $t$. It is mainly due to the local sensitivity of the solution in terms of $x$ and $t$. The periodical pattern in the $x$ axis allows GP to identify the actual location of the data more effectively. In contrast, the local sensitivity along the $t$ direction is close to zero, which provides limited additional information to refine the posterior. The means of the posterior also indicate the improvements in the location of uncertain data. In most cases, the posterior means are much closer to the actual location after the Bayesian update compared to their prior means.

The predictions of GP based on both certain and uncertain data are shown in Figure~\ref{fig:HE_GP_prediction}. The mean prediction of GP with the posterior distribution of the uncertain data demonstrates substantially lower absolute errors and variances compared to those based on the prior distribution. It indicates the effectiveness of the Bayesian update to the uncertain information in the GP training process.

\subsection{Reaction-diffusion equation}
\begin{figure}[th!]
    \centering
    \includegraphics[width=\linewidth]{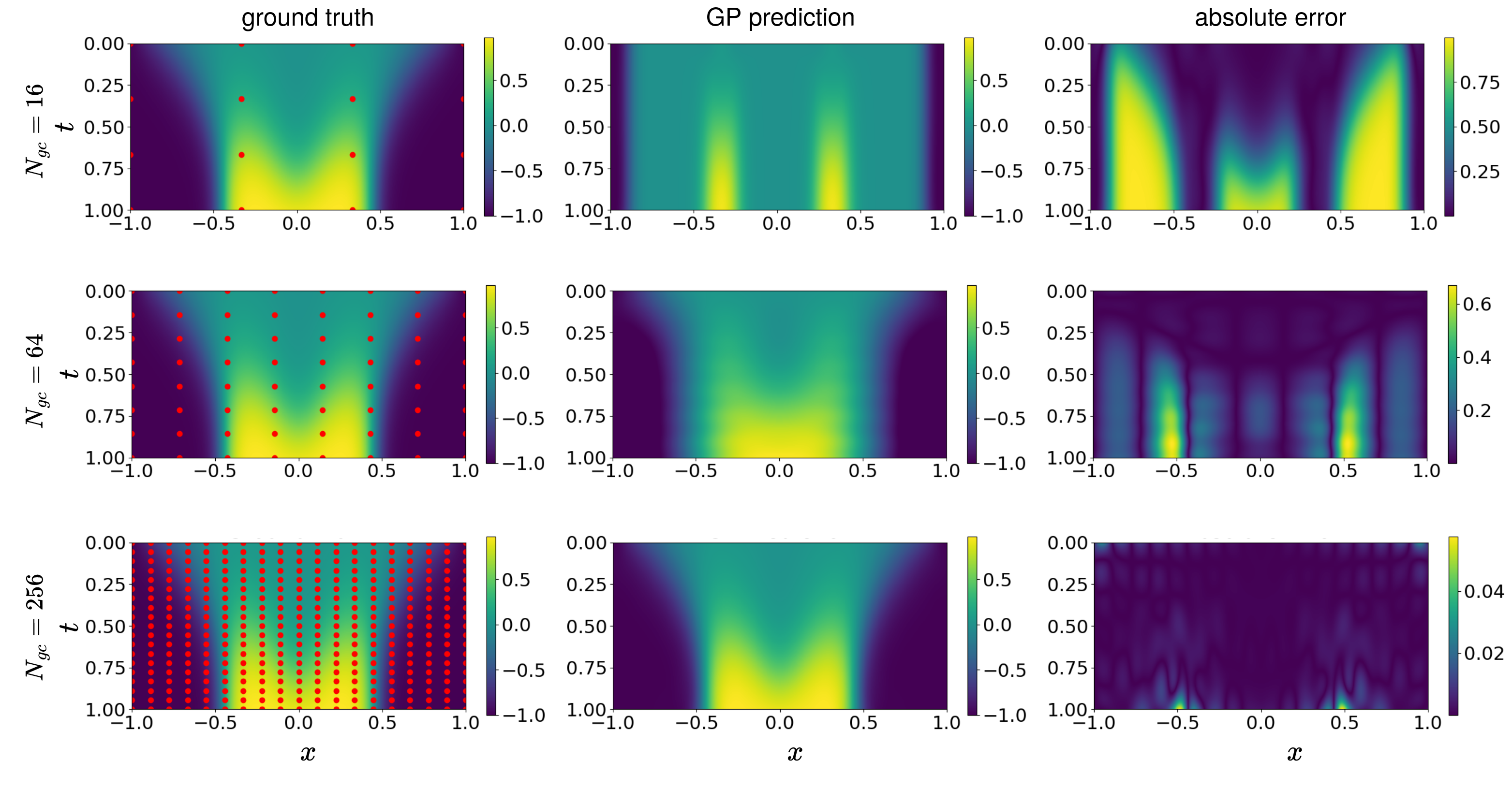}
    \caption{Comparisons of GP prediction and ground truth with training sample size $N_{gc}= 16$, 64 and 256 for the source term. Their corresponding absolute errors are also shown. The red dots on the first column represent the locations of training data of the source term.}
    \label{fig:RD_learning}
\end{figure}

In this experiment, a reaction-diffusion equation example is demonstrated. The reaction-diffusion model was originally used for simulating chemical reactions and has been extended to a variety of disciplines such as biology, physics and ecology for other dynamical processes \cite{allen1975coherent,tersian2001periodic,kondo2010reaction}. In this example, we present a one-dimensional Allen-Cahn reaction-diffusion equation \cite{allen1975coherent},

\begin{equation}
\begin{dcases}
  \frac{\partial z({x},t)}{\partial t} - \alpha\frac{\partial^2 z({x},t)}{\partial x^2} +\beta(z^3-z) = 0,    & (x,t) \in (-1,1) \times (0,1],\\
  z(-1,t) =z(1,t), \\
  \frac{\partial z(-1,t)}{\partial x} = \frac{\partial z(1,t)}{\partial x},\\
   z(x,0) = x^2\cos{\pi x} ,
\end{dcases} 
\end{equation}\\
where the coefficients $\alpha$ and $\beta$ are set to be 0.01 and 5 respectively. It is important to note that with the reaction term $R(z) = \beta(z^3 -z)$, the problem is no longer a linear PDE. However a joint distribution can still be formed between $z$ and $\mathcal{L}z$, where $\mathcal{L} = (\frac{\partial}{\partial t} - \frac{\partial^2}{\partial x^2})$ is the linear differential operator. The major difference lies in data generation. In the heat equation example, the data of $\mathcal{L}z$ are generated from source term $g(x,t)$, while here are generated from $ - R(z)$. 

\begin{figure}[t!]
    \centering
    \includegraphics[width=\linewidth]{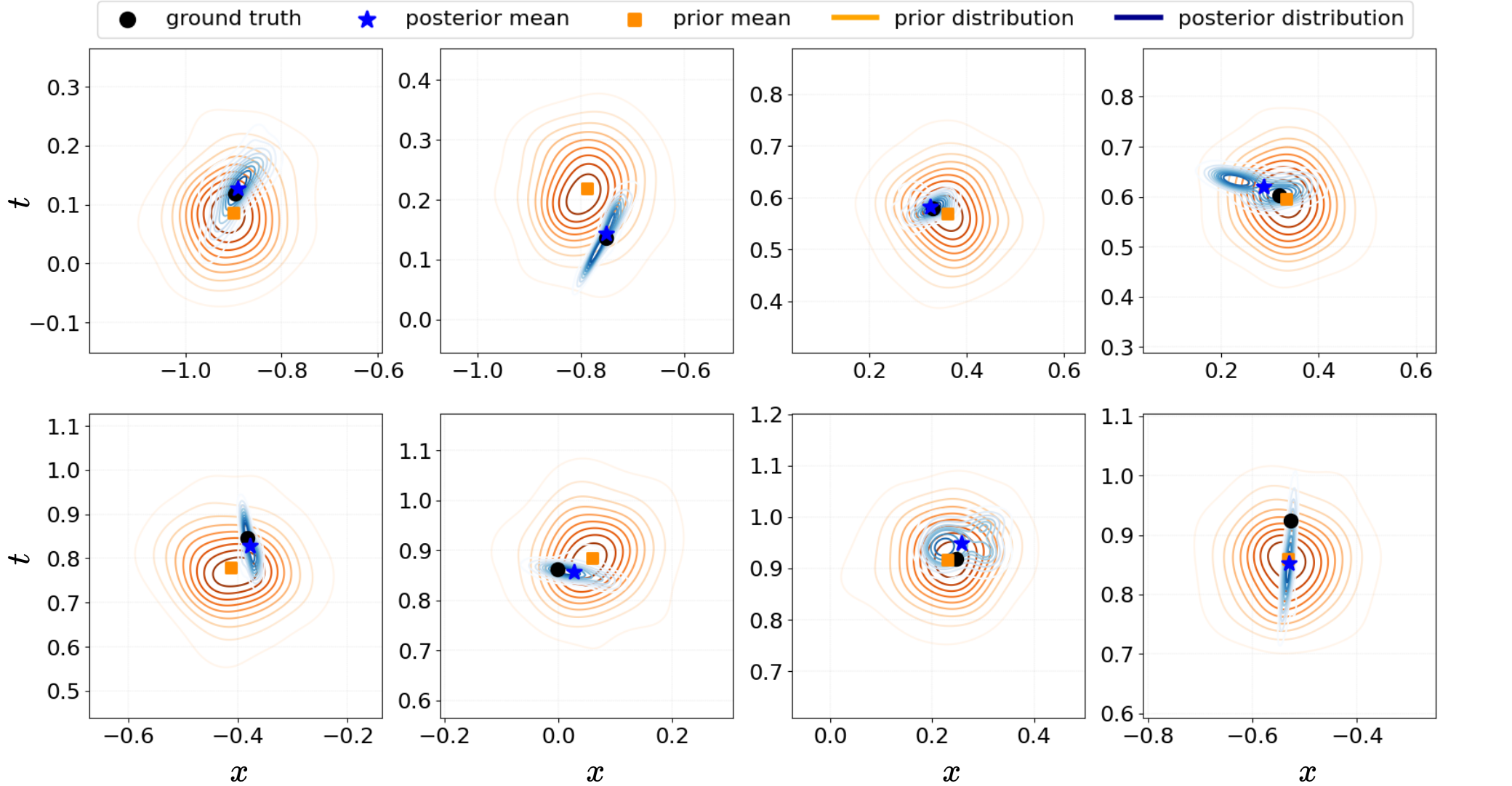}
    \caption{Prior and posterior distributions of eight data with uncertainties in location and time. The black dots denote the actual locations of data. Orange contours and squares denote their prior distributions and means. Blue contours and stars denote the posterior distributions and means.}
    \label{fig:RD_post}
\end{figure}

\begin{figure}[t!]
    \centering
    \includegraphics[width=\linewidth]{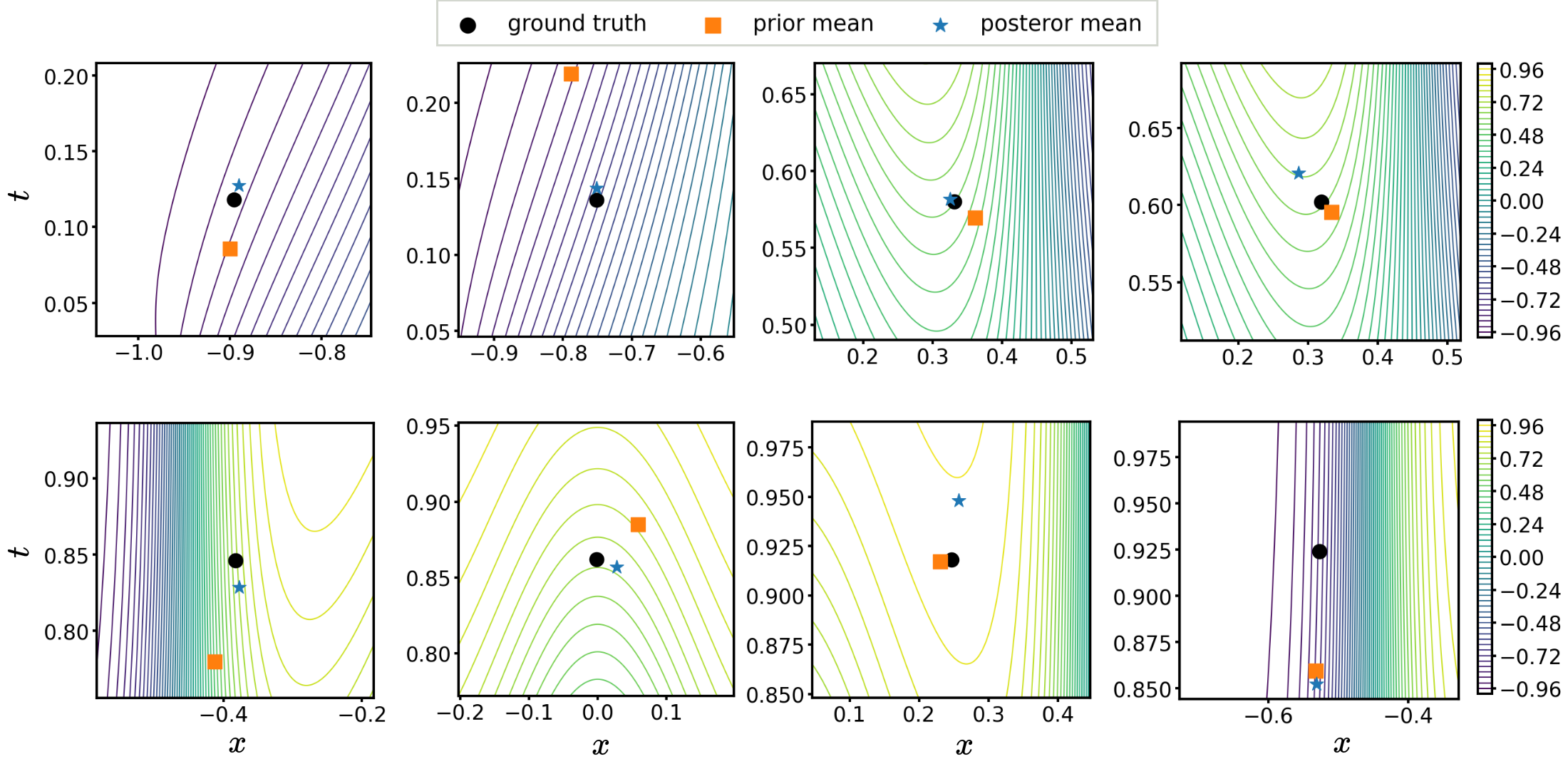}
    \caption{Prior and posterior mean of eight uncertain data and their corresponding ground truth locations. The contour lines in the background represent the contour of solution values. The black dots denote the actual locations of data. Orange squares denote their prior means, while blue stars denote the posterior means.}
    \label{fig:RD_coutour}
\end{figure}

\begin{figure}[th!]
    \centering
    \includegraphics[width=\linewidth]{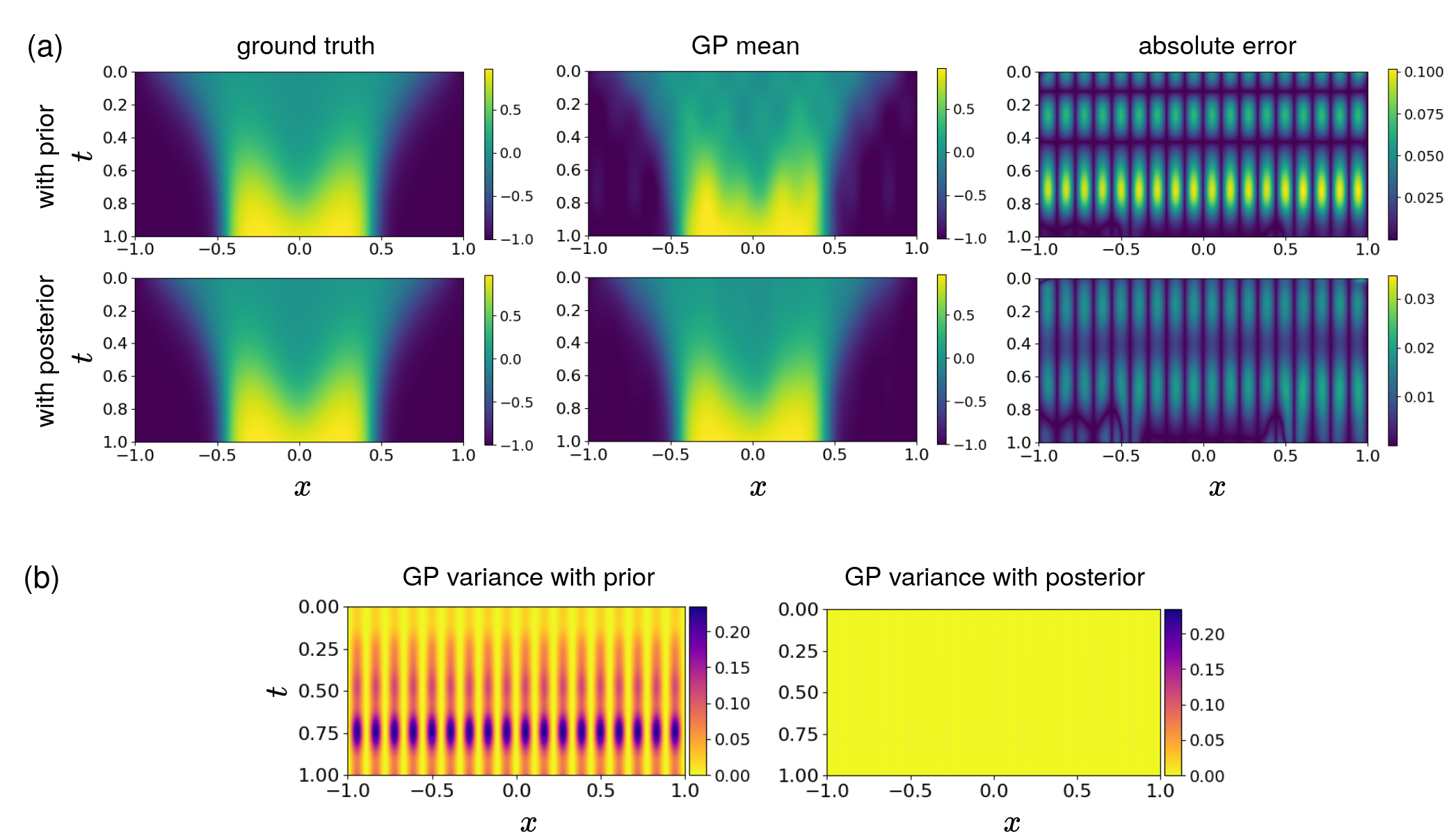}
    \caption{Means and variances of GP predictions for reaction-diffusion equation combining with prior and posterior distributions of uncertainty data locations.}
    \label{fig:RD_pred}
\end{figure}

Different from the heat equation example, evenly spaced samples are generated to capture the phase separation dynamics of Allen-Cahn equation, as shown in Figure~\ref{fig:RD_learning}. It can be observed that GP emulate the solution of PDE sufficiently well when the number of samples reaches 256. The maximum absolute error is reduced to around 0.055. 

Eight uncertain locations are presumed in this case with respect to both $t$ and $x$. Similar to the previous example, the prior means of these data are generated from their actual locations corrupted by a noise that follows a normal distribution $\mathcal{N}(0,0.04^2)$ and the prior variance is set to be 0.04. The marginal posterior distributions of those data after inference are demonstrated in Figure~\ref{fig:RD_post}. The majority of the uncertain locations have been significantly improved, for example, the first three data locations in the upper row of Figure~\ref{fig:RD_post}. The posterior means of the data represented by blue stars almost overlap with their corresponding ground truth represented by black circles. Clear improvement can also be observed from the first two data points in the lower row of Figure~\ref{fig:RD_post}. The rest of the data points merely show any improvement. This is mainly due to the local sensitivity of the solution at different locations. A visualization of the prior and posterior means of the uncertain data locations as well as their actual location is presented in Figure~\ref{fig:RD_coutour} with a background of the solution contour lines. When the local sensitivity is close to zero, it indicates that the prior uncertain location has a similar solution value as its actual location. For example, the prior means of the two data locations in the lower row of Figure~\ref{fig:RD_coutour} spot at the same contour area as their ground truth. Therefore it is hard for the GP model to provide any additional useful information to correct its mean and reduce the corresponding variance. 

Figure~\ref{fig:RD_pred} exhibits the mean and variance of GP predictions for the entire domain based on both certain and uncertain data. The results with prior and posterior are demonstrated in Figure~\ref{fig:RD_pred}(a) respectively. The result with posterior distribution of the uncertain data again demonstrates substantially lower errors and less variance compared to those with their prior distributions.

\section{Conclusion}\label{sec:conclusion}
In this work, a Bayesian method that integrates uncertain data locations into Gaussian process regression is proposed and demonstrated. Through Bayesian inference, all available information from the data is leveraged to update the knowledge on uncertain input locations. By marginalizing the posterior distribution of the input uncertainties, the predictive distribution for new test points can be achieved via Gaussian process regression. The method can be applied to the scenario of constructing a surrogate model for PDEs with GP in order to reduce the uncertainties in the solution data inherited from measurement. Multiple numerical experiments have demonstrated that, in comparison with the results without Bayesian updating of data locations, the proposed method presents a consistently improved performance in generalization and predictive uncertainty reduction for both one-dimensional functions and PDE scenarios.

\section*{Funding}
D. Ye and M. Guo acknowledge the financial support from Sectorplan Bèta (NL) under the focus area \emph{Mathematics of Computational Science}. D. Ye also acknowledges the financial support from NGF AiNed XS Europe under No. NGF.1609.241.020. W. Yan acknowledges the China Scholarship Council for their support under No. 202107650017. M. Guo and C. Brune also acknowledge the support from the 4TU Applied Mathematics Institute for the Strategic Research Initiative \emph{Bridging Numerical Analysis and Machine Learning}. 

\section*{Data availability}
All the data and source codes to reproduce the results in this study are available on GitHub at \url{https://github.com/DongweiYe/GP-linearPDE-uncertainty}.

\bibliographystyle{plain}
\bibliography{reference}

\begin{thebibliography}{10}

\bibitem{Alaa2017}
Ahmed~M. Alaa and Mihaela van~der Schaar.
\newblock {B}ayesian inference of individualized treatment effects using
  multi-task {G}aussian processes.
\newblock In I.~Guyon, U.~Von Luxburg, S.~Bengio, H.~Wallach, R.~Fergus,
  S.~Vishwanathan, and R.~Garnett, editors, {\em Advances in Neural Information
  Processing Systems}, volume~30, 2017.

\bibitem{allen1975coherent}
Samuel~M Allen and John~W Cahn.
\newblock Coherent and incoherent equilibria in iron-rich iron-aluminum alloys.
\newblock {\em Acta Metallurgica}, 23(9):1017--1026, 1975.

\bibitem{alvarez2009latent}
Mauricio Alvarez, David Luengo, and Neil~D Lawrence.
\newblock Latent force models.
\newblock In {\em Artificial intelligence and statistics}, pages 9--16. PMLR,
  2009.

\bibitem{arzani2021data}
Amirhossein Arzani and Scott~TM Dawson.
\newblock Data-driven cardiovascular flow modelling: examples and
  opportunities.
\newblock {\em Journal of the Royal Society Interface}, 18(175):20200802, 2021.

\bibitem{batlle2023error}
Pau Batlle, Yifan Chen, Bamdad Hosseini, Houman Owhadi, and Andrew~M Stuart.
\newblock Error analysis of kernel/gp methods for nonlinear and parametric
  pdes.
\newblock {\em arXiv preprint arXiv:2305.04962}, 2023.

\bibitem{besginow2022constraining}
Andreas Besginow and Markus Lange-Hegermann.
\newblock Constraining gaussian processes to systems of linear ordinary
  differential equations.
\newblock {\em Advances in Neural Information Processing Systems},
  35:29386--29399, 2022.

\bibitem{bock2019review}
Frederic~E Bock, Roland~C Aydin, Christian~J Cyron, Norbert Huber, Surya~R
  Kalidindi, and Benjamin Klusemann.
\newblock A review of the application of machine learning and data mining
  approaches in continuum materials mechanics.
\newblock {\em Frontiers in Materials}, 6:110, 2019.

\bibitem{brunton2024promising}
Steven~L Brunton and J~Nathan Kutz.
\newblock Promising directions of machine learning for partial differential
  equations.
\newblock {\em Nature Computational Science}, pages 1--12, 2024.

\bibitem{brunton2020machine}
Steven~L Brunton, Bernd~R Noack, and Petros Koumoutsakos.
\newblock Machine learning for fluid mechanics.
\newblock {\em Annual Review of Fluid Mechanics}, 52(1):477--508, 2020.

\bibitem{BURKHART2014189}
Michael~C. Burkhart, Yeonsook Heo, and Victor~M. Zavala.
\newblock Measurement and verification of building systems under uncertain
  data: A {G}aussian process modeling approach.
\newblock {\em Energy and Buildings}, 75:189--198, 2014.

\bibitem{Chang2015}
Eugene T~Y Chang, Mark Strong, and Richard~H Clayton.
\newblock {B}ayesian sensitivity analysis of a cardiac cell model using a
  {G}aussian process emulator.
\newblock {\em PLOS ONE}, 10(6):1--20, 06 2015.

\bibitem{chen2021solving}
Yifan Chen, Bamdad Hosseini, Houman Owhadi, and Andrew~M Stuart.
\newblock Solving and learning nonlinear pdes with gaussian processes.
\newblock {\em Journal of Computational Physics}, 447:110668, 2021.

\bibitem{costabal2019multi}
Francisco~Sahli Costabal, Paris Perdikaris, Ellen Kuhl, and Daniel~E Hurtado.
\newblock Multi-fidelity classification using {G}aussian processes:
  accelerating the prediction of large-scale computational models.
\newblock {\em Computer Methods in Applied Mechanics and Engineering},
  357:112602, 2019.

\bibitem{Cuomo2022}
Salvatore Cuomo, Vincenzo~Schiano Di~Cola, Fabio Giampaolo, Gianluigi Rozza,
  Maziar Raissi, and Francesco Piccialli.
\newblock Scientific machine learning through physics--informed neural
  networks: Where we are and what's next.
\newblock {\em Journal of Scientific Computing}, 92(3):88, Jul 2022.

\bibitem{Dallaire2009}
Patrick Dallaire, Camille Besse, and Brahim Chaib-draa.
\newblock Learning {G}aussian process models from uncertain data.
\newblock In Chi~Sing Leung, Minho Lee, and Jonathan~H. Chan, editors, {\em
  Neural Information Processing}, pages 433--440, Berlin, Heidelberg, 2009.
  Springer Berlin Heidelberg.

\bibitem{DALLAIRE20111945}
Patrick Dallaire, Camille Besse, and Brahim Chaib-draa.
\newblock An approximate inference with {G}aussian process to latent functions
  from uncertain data.
\newblock {\em Neurocomputing}, 74(11):1945--1955, 2011.

\bibitem{Frigola2013}
Roger Frigola, Fredrik Lindsten, Thomas~B Sch\"{o}n, and Carl~Edward Rasmussen.
\newblock {B}ayesian inference and learning in {G}aussian process state-space
  models with particle {MCMC}.
\newblock In C.J. Burges, L.~Bottou, M.~Welling, Z.~Ghahramani, and K.Q.
  Weinberger, editors, {\em Advances in Neural Information Processing Systems},
  volume~26, 2013.

\bibitem{Ghaffari2007}
Maani Ghaffari~Jadidi, Jaime~Valls Miro, and Gamini Dissanayake.
\newblock Warped {G}aussian processes occupancy mapping with uncertain inputs.
\newblock {\em IEEE Robotics and Automation Letters}, 2(2):680--687, 2017.

\bibitem{Girard2004}
Agathe Girard.
\newblock Approximate methods for propagation of uncertainty with {G}aussian
  process models.
\newblock {\em PQDT - Global}, page 168, 2004.

\bibitem{Gulian2019}
Mamikon Gulian, Maziar Raissi, Paris Perdikaris, and George Karniadakis.
\newblock Machine learning of space-fractional differential equations.
\newblock {\em SIAM Journal on Scientific Computing}, 41(4):A2485--A2509, 2019.

\bibitem{kondo2010reaction}
Shigeru Kondo and Takashi Miura.
\newblock Reaction-diffusion model as a framework for understanding biological
  pattern formation.
\newblock {\em Science}, 329(5999):1616--1620, 2010.

\bibitem{Krige1951}
D.G. Krige.
\newblock A statistical approach to some basic mine valuation problems on the
  witwatersrand.
\newblock {\em Journal of the Southern African Institute of Mining and
  Metallurgy}, 52(6):119--139, 1951.

\bibitem{Mchutchon2011}
Andrew Mchutchon and Carl Rasmussen.
\newblock {G}aussian process training with input noise.
\newblock In J.~Shawe-Taylor, R.~Zemel, P.~Bartlett, F.~Pereira, and K.Q.
  Weinberger, editors, {\em Advances in Neural Information Processing Systems},
  volume~24, 2011.

\bibitem{phellan2021real}
Renzo Phellan, Bahe Hachem, Julien Clin, Jean-Marc Mac-Thiong, and Luc Duong.
\newblock Real-time biomechanics using the finite element method and machine
  learning: Review and perspective.
\newblock {\em Medical Physics}, 48(1):7--18, 2021.

\bibitem{Quionero2003}
Joaquin Quiñonero-Candela and Sam~T Roweis.
\newblock Data imputation and robust training with {G}aussian processes.
\newblock NIPS, 2003.

\bibitem{raissi2018hidden}
Maziar Raissi and George~Em Karniadakis.
\newblock Hidden physics models: Machine learning of nonlinear partial
  differential equations.
\newblock {\em Journal of Computational Physics}, 357:125--141, 2018.

\bibitem{raissi2018numerical}
Maziar Raissi, Paris Perdikaris, and George~Em Karniadakis.
\newblock Numerical gaussian processes for time-dependent and nonlinear partial
  differential equations.
\newblock {\em SIAM Journal on Scientific Computing}, 40(1):A172--A198, 2018.

\bibitem{williams2006gaussian}
Carl~E Rasmussen and Christopher K~I Williams.
\newblock {\em {G}aussian Processes for Machine Learning}.
\newblock The MIT Press, 2006.

\bibitem{Mats1982}
Mats Rudemo.
\newblock Empirical choice of histograms and kernel density estimators.
\newblock {\em Scandinavian Journal of Statistics}, 9(2):65--78, 1982.

\bibitem{sanderse2024scientific}
Benjamin Sanderse, Panos Stinis, Romit Maulik, and Shady~E Ahmed.
\newblock Scientific machine learning for closure models in multiscale
  problems: A review.
\newblock {\em arXiv preprint arXiv:2403.02913}, 2024.

\bibitem{sarkka2011linear}
Simo S{\"a}rkk{\"a}.
\newblock Linear operators and stochastic partial differential equations in
  gaussian process regression.
\newblock In {\em Artificial Neural Networks and Machine Learning--ICANN 2011:
  21st International Conference on Artificial Neural Networks, Espoo, Finland,
  June 14-17, 2011, Proceedings, Part II 21}, pages 151--158. Springer, 2011.

\bibitem{Sobol1999}
I.~M. Sobol and Yu.~L. Levitan.
\newblock A pseudo-random number generator for personal computers.
\newblock {\em Computers and Mathematics with Applications}, 37(4-5):33--40,
  1999.

\bibitem{tersian2001periodic}
Stepan Tersian and Julia Chaparova.
\newblock Periodic and homoclinic solutions of extended fisher--kolmogorov
  equations.
\newblock {\em Journal of Mathematical Analysis and Applications},
  260(2):490--506, 2001.

\bibitem{ye2024gaussian}
Dongwei Ye and Mengwu Guo.
\newblock Gaussian process learning of nonlinear dynamics.
\newblock {\em Communications in Nonlinear Science and Numerical Simulation},
  page 108184, 2024.

\bibitem{Ye2022}
Dongwei Ye, Pavel Zun, Valeria Krzhizhanovskaya, and Alfons~G. Hoekstra.
\newblock Uncertainty quantification of a three-dimensional in-stent restenosis
  model with surrogate modelling.
\newblock {\em Journal of The Royal Society Interface}, 19(187):20210864, 2022.

\end{thebibliography}

\end{document}